\documentclass[10pt,twocolumn,letterpaper]{article}

\usepackage[pagenumbers]{cvpr} 

\newcommand{\tit}[1]{\smallbreak\noindent\textbf{#1.}}

\newcommand{\token}[1]{$<\!\!#1\!\!>$}
\DeclareMathOperator*{\argmax}{arg\,max}

\DeclareMathOperator*{\append}{append}

\newcommand{\method}{ToFu\xspace}
\newcommand{\dataset}{ComPairs\xspace}

\usepackage{algorithmic}
\usepackage{algorithm}
\usepackage{graphicx}
\usepackage{amssymb}
\usepackage{pifont}
\usepackage{makecell}
\usepackage{tabularx}
\usepackage{multirow}
\usepackage[dvipsnames,table]{xcolor}
\usepackage{balance}

\newcommand{\cmark}{\textcolor{ForestGreen}{\ding{51}}}%
\newcommand{\xmark}{\textcolor{BrickRed}{\ding{55}}}%
\definecolor{cvprblue}{rgb}{0.21,0.49,0.74}
\usepackage[pagebackref,breaklinks,colorlinks,allcolors=cvprblue]{hyperref}

\title{ToFu: Visual Tokens Reduction via Fusion \\ for Multi-modal, Multi-patch, Multi-image Tasks}

\author{
Vittorio Pippi$^{1\thanks{The main work was done while interning at Amazon.}}$ \qquad Matthieu Guillaumin $^2$ \qquad Silvia Cascianelli $^1$ \qquad Rita Cucchiara$^1$ \\ Maximilian Jaritz$^2$ \qquad Loris Bazzani$^{2}$\\\\
\begin{tabular}{ccc}
\makecell{$^1$University of Modena and Reggio Emilia} & \makecell{$^2$Amazon}\\
\end{tabular}
}

\begin{document}
\maketitle
\begin{abstract}
Large Multimodal Models (LMMs) are powerful tools that are capable of reasoning and understanding multimodal information beyond text and language. 
Despite their entrenched impact, the development of LMMs is hindered by the higher computational requirements compared to their unimodal counterparts. 
One of the main causes of this is the large amount of tokens needed to encode the visual input, which is especially evident for multi-image multimodal tasks. 
Recent approaches to reduce visual tokens depend on the visual encoder architecture, require fine-tuning the LLM to maintain the performance, and only consider single-image scenarios. 
To address these limitations, we propose \method, a visual encoder-agnostic, training-free Token Fusion strategy that combines redundant visual tokens of LMMs for high-resolution, multi-image, tasks. 
The core intuition behind our method is straightforward yet effective: preserve distinctive tokens while combining similar ones. 
We achieve this by sequentially examining visual tokens and deciding whether to merge them with others or keep them as separate entities.
We validate our approach on the well-established LLaVA-Interleave Bench, which covers challenging multi-image tasks. 
In addition, we push to the extreme our method by testing it on a newly-created benchmark, \dataset, focused on multi-image comparisons where a larger amount of images and visual tokens are inputted to the LMMs. 
Our extensive analysis, considering several LMM architectures, demonstrates the benefits of our approach both in terms of efficiency and performance gain.
\end{abstract}    
\section{Introduction}
\label{sec:intro}

\begin{figure}
    \centering
    \includegraphics[width=\linewidth]{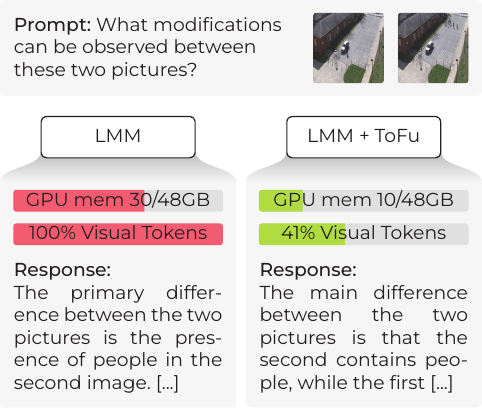}
    \caption{Our proposed \method allows reducing the number of visual tokens in input to the LLM in LMMs while preserving the same information. This allows reducing runtime and GPU memory consmption without affecting the performance. The image refers to the InternVL2 4B model and the GPU NVIDIA L40S with 48GB of memory.}
    \label{fig:overview}
\end{figure}

Recent years have witnessed the widespread of Large Language Models (LLMs), able to solve a number of different language-related tasks, and, more recently, LMMs~\cite{chen2024internvl, zhu2024minigpt,liu2023llava,liu2023improvedllava,chen2024far,jiang2024mantis, wang2024qwen2, ye2024mplug, mckinzie2024mm1, wu2023next, dubey2024llama, agrawal2024pixtral, openai2024gpt4,geminiteam2024gemini,anthropic2024claude}, which bridge the visual and textual modalities. 
LMMs rely on a rich, multimodal representation of concepts, which is built upon both visual and textual tokens, that are often aligned in a shared embedding space. 
Nonetheless, the development of LMMs is slower than that of their language counterparts, especially in scenarios involving a large number of images, \eg, multi-image question answering~\cite{tanaka2023slidevqa,bansal2020visual}, multi-view reasoning~\cite{hong20233d}, and video understanding~\cite{li2023videochat,buch2022revisiting}. 
One of the reasons for this is the complexity of aligning and integrating multiple modalities, both from a technical point of view and for computational and memory requirements limitations.
Note that the high computational requirements of LMMs are the main aspects that potentially reduce their applicability in scenarios in which computational resources are limited. In this respect, strategies for decreasing the memory footprint of LMMs contribute to making these models more sustainable, \ie, scalable and usable by a larger user base. This consideration inspired recent works to focus on LMMs efficiency.

Most LMMs consist of two main components: a LLM for reasoning and language generation, and a visual encoder that extracts visual tokens from input images to enrich the context for the LLM. 
The computational demands of LMMs are predominantly attributed to the LLM, which is usually more extensive than the visual encoder. 
This computational cost stems not only from the LLM's vast number of parameters and operations but also from the length of the context it processes, which incurs quadratic complexity in the Transformer's attention mechanisms.
In multi-image tasks, the number of image tokens serving as context significantly outnumbers the language tokens, thus substantially increasing the inference cost of LMM in terms of latency and memory footprint. As an example, for the LMMs in the InternVL2 family~\cite{chen2024internvl}, the visual tokens are $\sim$98.66\% of the total number of tokens consumed by the featured LLM. 

In order to overcome these limitations, we argue that it is important to take into consideration that text and images are inherently very different vehicles of semantic concepts, which is reflected in the characteristics of their token-based representation. Specifically, a piece of text can be represented compactly and efficiently with a relatively small set of discrete tokens (\eg, subwords). On the other hand, when representing an image with a set of visual tokens, visual concepts can be represented multiple times by similar tokens, which can be redundant. 
This is in line with findings of recent works~\cite{bolya2023token,liu2023adaptive} that have demonstrated that many visual tokens extracted by the visual encoders used in LMMs do not significantly contribute to the output computation, and thus, can be disregarded with minimal impact on the performance. 

LMMs increasingly face challenges in handling complex visual inputs, which generate a substantial number of visual tokens. 
These inputs encompass high-resolution images (multi-patch), videos (multi-frame), 3D scenes (multi-view), and image sets (multi-image)~\cite{li2024llavaNI}.
While previous optimization strategies have focused on reducing LMM size through smaller architectures~\cite{chu2023mobilevlm,yuan2024tinygpt}, model quantization~\cite{ma2024era,xie2024advancing}, and weight pruning~\cite{wang2024large}, these approaches often compromise reasoning capabilities. 
Notably, there has been insufficient attention given to reducing redundant visual tokens~\cite{kitaev2019reformer,wang2020linformer,bolya2023token,chen2024image,li2024flexattention,zhang2024token,arif2024hired,shang2024llavaPM}, despite its potential for significant efficiency gains.
To address this gap, we propose \method, a training-free, visual encoder-agnostic solution that reduces the number of prefix visual tokens fed to the LLM.
\method adapts to the number of images involved, determining the optimal quantity of visual tokens to retain.
Our approach involves computing a new sequence of tokens from the original visual token sequence, minimizing redundancy. 
The core intuition behind our method is straightforward yet effective: preserve distinctive tokens while combining similar ones. 
We achieve this by sequentially examining visual tokens and deciding whether to merge them with others or keep them as separate entities.
This method is particularly effective for multi-patch and multi-image tasks, where visual token redundancy is common. 
By reducing the token count, we enhance efficiency without sacrificing the model's ability to process complex visual information.

To evaluate the proposed approach, we consider a large set of different multimodal, multi-image tasks, including the well-established LLaVA-Interleave Benchmark~\cite{liu2023improvedllava}. 
It is composed of nine in-domain evaluation datasets that entail tasks such as visual question answering, spotting the difference, visual storytelling, multi-image puzzles, visual reasoning, image sequence completion, visual quality assessment, and image editing instruction generation. In all cases, the visual input consists of multiple single-patch images, in most cases between two and six.
In addition, we push to the extreme our method by testing it on a newly-created benchmark, \dataset, focused on multi-image comparisons where a larger amount of images and visual tokens are inputted to the LMMs. In fact, the samples in ComPairs have on average more than 14.6k visual tokens, which is one order of magnitude more than the average number of visual tokens per sample in the datasets featured in LLaVA-Interleave Benchmark ($\sim$4.8k tokens on average).
The samples in the dataset feature two sets of multiple high-resolution images of similar objects, and the LMM is tasked to compare the two objects based on the provided image sets. 
These characteristics make ComPairs particularly challenging, also with respect to the datasets in LLaVA-Interleave Bench.
Extensive evaluation on the considered datasets and in combination with different open-source LLMs demonstrates the effectiveness of our approach, which is able to reduce the number of tokens (and thus, save computational resources) up to 60\% without negatively affecting the performance. 

\section{Related Work}

In literature, a number of techniques were proposed to reduce the computational requirements of LLMs and LMMs, which involve distillation of larger architectures~\cite{chen2024internvl}, training smaller models with higher-quality data~\cite{zhou2024tinyllavaframeworksmallscalelarge}, or more optimized modules and training~\cite{zhang2024tinyllamaopensourcesmalllanguage, chu2023mobilevlmfaststrong}.
Our analysis here focuses on the reduction of visual tokens in LMMs, specifically useful for high-resolutions images and multi-image, multi-patch inputs.
Most work in this domain focuses on reduction within the visual encoder, within the LLM, or, as in our case, after the visual encoder but before the projection layer and the LLM, referred to as \emph{early dropping}.

\tit{Token Reduction within the Visual Encoder} 
When computing the visual tokens with a Transformer-like encoder, sparse attention can be implemented by limiting attention computation to localized regions, thus reducing the quadratic complexity of the attention operation~\cite{kitaev2019reformer,wang2020linformer}. Alternatively, attention can be computed on the full context, but tokens can be reduced at each in each transformer block, \eg, by performing bipartite matching to select the most representative tokens to be passed to the next block as is~\cite{bolya2023token}. Note that these strategies heavily depend on the visual encoder implementation and, therefore, are not straightforward to adapt to different architectures.

\tit{Token Reduction within the LLM} 
During the LLM decoding, visual tokens with low attention scores can be disregarded in some attention operations. For example, such less important tokens can be identified in the initial layers of the LLM and then skipped when performing attention in the following layers as in~\cite{chen2024image,li2024flexattention}. Note that, although this strategy allows limiting the computation requirements of non-initial layers in the LLM, it still entails processing all the tokens in the initial layers, thus only partially improving the overall model efficiency.

\tit{Token Reduction via Early Dropping}
A more efficient strategy entails reducing the number of visual tokens once extracted by the visual encoder, allowing only relevant tokens to be fed to the LLM. 
This approach is independent of the encoder’s architecture and reduces the amount of tokens the LLM has to deal with. 
This strategy can be implemented by dropping repetitive tokens via cosine similarity, as in~\cite{zhang2024token}, which removes white spaces in visual inputs for document understanding tasks. 
Recent methods~\cite{zhang2024token,shang2024llavaPM} leverage on the self-attention score between the spatial visual tokens and the CLS token to identify relevant tokens. 
HiRED~\cite{arif2024hired} focuses on dropping redundant tokens for tasks involving single high-resolution images (treated as a multi-patch input to the LMM). 
PruMerge~\cite{shang2024llavaPM} relies on CLS-attention to identify the most relevant spatial visual tokens and combines them with k of the other tokens, based on their relative self-attention. To prevent excessive information loss due to pruning PruMerge+~\cite{shang2024llavaPM} also includes uniformly sampled visual tokens in the context for the LLM. \method is also in this category of token reduction via early dropping. However, while previous works are designed for single-image tasks, we focus on multi-image tasks. Moreover, our proposed solution is intended to be task-agnostic and easily adaptable to any visual encoder implementation. 

\tit{Datasets for Token Reduction}
There exist several datasets to train and benchmark LMMs~\cite{liu2023improvedllava, Li_2024_CVPR, liu2025mmbench, fu2024mmecomprehensiveevaluationbenchmark}.
SEED-Bench~\cite{Li_2024_CVPR} is designed to evaluate hierarchical capabilities of LMMs, while structuring the task in multiple levels.
MME~\cite{fu2024mmecomprehensiveevaluationbenchmark} covers the examination of perception and cognition abilities.
MMBench~\cite{liu2025mmbench} focuses on multi-lingual and designs quality control schemes to increase accuracy of evaluation.
Among these, LLaVA-Interleave Bench~\cite{liu2023improvedllava} is one of the most comprehensive datasets that also focuses on multi-image sets.
For this reasons, we use LLaVA-Interleave Bench in our work to test the capabilities of token reduction methods.
In addition, we propose ComPairs which contains higher resolutions images and an average number of visual tokens per sample of 14K against 3-8K of LLaVA-Interleave Bench.

\section{Proposed Approach}
\label{sec:tofu}
A vast majority of modern LMMs~\cite{liu2023llava, chen2024internvl, jiang2024mantis} are designed with the following structure: images are fed into a Visual Encoder that extracts visual embeddings, an Adapter connects these embeddings into the textual embedding space, and an LLM consumes both visual and textual embedding tokens to generate its textual output. 
The proposed algorithm aims at reducing the number of visual embedding tokens that are outputted by the Adapter, which we will call \emph{visual tokens} for brevity in the rest of the paper.
Performing reduction of visual tokens at this stage of the LMM pipeline is deemed to be the most beneficial in terms of efficiency gain. 
In fact, the visual and textual input tokens are injected in the LLM every time that it generates the next token during inference, while the Visual Encoder and the Adapter are called just once at the beginning of the process.
Since the next token computation has quadratic complexity with respect to the number of LLM input tokens, reducing the number of tokens can significantly impact efficiency in terms of speed and memory. 
Given the sequence of visual tokens, \method entails computing a new sequence of tokens containing as least redundant elements as possible. 
The main intuition to design \method is to keep tokens that are \emph{distinctive} while combining tokens that are \emph{similar} to each other. 

\method takes as input a sequence of $M$ visual tokens $V=[v_1,...,v_{M}]$, where $v_m\in\mathbb{R}^N$ is the $m$-th visual token in the sequence of dimensionality $N$.
The objective is to create a new target sequence of visual tokens $T=[t_1,...,t_{K}]$ that contains a set of distinctive tokens with the constraint that $K << M$.
The proposed method iteratively inspects the visual tokens in $V$ and decides if 1) they are worth to be kept in $T$ because of their distinctiveness or 2) if they should be combined with an element already in $T$.
Distinctiveness is defined here as checking if the similarity of two visual tokens exceeds a certain threshold $\tau$.
Moreover, the combination of visual tokens is done by taking into account their importance weight, therefore \method also keeps track of a list of weights $W$ that corresponds to visual tokens.

The algorithm starts by inserting the first element of the input $V$ in the target sequence, \ie, $T=[v_1]$, and setting its importance weight to $W=[1]$. 
For each of the following tokens in the input sequence, $v_m \in V$, we find the token $t_j \in T$ with the highest similarity with respect to $v_m$:
\begin{align*}
t_j = \argmax_{t_k \in T}(s(v_m, t_k))
\end{align*}
where $s(\cdot, \cdot)$ is defined as the Cosine Similarity between embeddings. 
If the similarity is lower than the threshold $s(v_m, t_j) \le \tau$, and thus, the $v_m$ token is a distinct new element, then we add it to the output $T=[t_1, ..., t_k, v_m]$, and update the weights accordingly as $W=[w_1, ..., w_k, 1]$. 
If $s(v_m, t_j) > \tau$, and thus, the token is very similar to an existing one in the target sequence, then we combine the token $t_j$ with $v_m$ and update its corresponding weight $w_j$, as follows:
\begin{align*}
    t_j =&~(t_j \cdot w_j + v_m) / (w_j + 1) \\
    w_j =&~w_j + 1
\end{align*}
This procedure iterates until all the tokens in $V$ are consumend and $T$ is therefore the target sequence that will be used by the LLM to perform inference.
We report the pseudo-code in~\Cref{alg:fusto_algorithm}. 

As argued in~\cite{cai2024matryoshka}, the optimal amount of visual tokens varies based on the specific multi-image task at hand. This can be controlled via the threshold $\tau$, which can be set based on the number of visual tokens in input and the desired size or percentage of the target sequence. 
In light of the argument made in~\cite{cai2024matryoshka}, for our approach, we propose a strategy to automatically set the threshold $\tau$ based on the length of the input tokens. 
First, we argue that the higher the number of visual tokens $M$, the more likely there will be redundant tokens fed to the LLM, and thus, the threshold $\tau$ should be set lower so that more tokens can be combined together. On the other hand, when $M$ is smaller, the token reduction algorithm can be less aggressive and $\tau$ can be given a higher value, ensuring that more input tokens are maintained.
Then we observe that among the LMMs available to date, those using the largest number of visual tokens are the InternVL series models (from $256$ to $3328$ per image). Therefore, we set a lower bound for the value of $\tau=0.7$ when $M=3328$, and an upper bound of $\tau=0.9$ when $M=256$ and interpolate between these two bounds for automatically determining the $\tau$ at different values of $M$.

The \method algorithm entails processing the visual embeddings sequentially. 
This helps preserve the original ordering of the tokens as much as possible, thus, maintaining the performance of the underlying LLM, which is used to receive a specific ordering as input. 
Moreover, a sequential implementation, in which we compute the similarity between one input embedding at a time to the output embeddings in $T$, is relatively computationally inexpensive. The alternative would be to compute a similarity matrix between all the tokens in $V$ and those in $T$ to find the optimal pair of embeddings to fuse together. However, $T$ changes at every token insertion, causing to repeat the quadratic similarity computation operation from scratch. Since we empirically noticed that processing the tokens sequentially is a good approximation, we opt for this strategy for efficiency.

\begin{algorithm}[t]
\caption{\method visual token reduction strategy}
\label{alg:fusto_algorithm}
\begin{algorithmic}[1]
    \STATE \textbf{Input: $V = [v_1, ..., v_M]$}
    \STATE \textbf{Output:} $T = [t_1, ...,  t_K]$
    \STATE $\tau$ \hfill $\triangleright$ Set similarity threshold
    \STATE $T = [v_1]$
    \STATE $W = [1]$
    \FOR{$m = 2, ..., M$}
        \STATE $t_j = \argmax_{t_k \in T}(s(v_m, t_k))$
        \IF{ $s(v_m, t_j) \le \tau$}
            \STATE $T.\append(v_m)$
            \STATE $W.\append(1)$
        \ELSE
            \STATE $t_j = (t_j \cdot w_j+v_m)/(w_j+1)$
            \STATE $w_j = w_j + 1$
        \ENDIF
    \ENDFOR
    \RETURN $T$
\end{algorithmic}
\end{algorithm}
\section{ComPairs Benchmark}
\label{sec:compairs}

To stress-test the currently available LMMs and validate their capability to deal with multiple high-resolution images and with an out-of-domain task, we devise the ComPairs benchmark for multi-image sets comparison. 
The ComPairs benchmark is composed by a collection of pairs of products derived from the publicly available Amazon Berkeley Objects (ABO)~\cite{collins2022abo} dataset. 
ABO is large-scale dataset that contains product catalog images, metadata, and artist-created 3D models with complex geometries and physically-based materials that correspond to real, household objects.
In this work, we focus on the image data and textual metadata with the goal of composing a dataset with multiple images of pairs of products and a detailed textual description of the differences between the products.

The data generation pipeline is divided in the following steps: 1) create the pairs of products, and 2) generate the text for comparing the images of the products.
In the first step, we filter products for each product category with descriptions in English and with at least three images in order to have a significant number of images that the LMM will use to make the comparisons.
To make the comparison task challenging, we create pairs of similar products from the same category using the BERT similarity score~\cite{zhangbertscore} between their descriptions.
We use a BERT similarity score in the range $[0.7, 0.9]$, which create challenging pairs while ensuring to discard nearly-identical products. 

Once the list of pairs are created, we feed their textual descriptions to the Llama3-8B LLM~\cite{dubey2024llama} and task it to identify a variable number of characteristics of both products.
For example, when comparing two sets of socks, relevant characteristics could be their {color}, the {number of items} in the set, their {length}, and the {material}. 
Moreover, we prompted the LLM to briefly describe the differences between the two products in terms of each of the identified characteristics. 
An example is shown in Fig.~\ref{fig:sample}.
We will share the prompts and more examples in the Supplementary Material.

\begin{table}[t]
    \centering
    \setlength{\tabcolsep}{.18em}
    \resizebox{\columnwidth}{!}{%
    \begin{tabular}{l c cccc}
    \toprule
    & 
    & \textbf{\#Samples} 
    & \textbf{avg \#Imgs}
    & \textbf{avg \#Tokens/Sample}
    & \textbf{\#Tokens/Img}\\
    \midrule
    \textbf{Spot the Difference}    && ~300	& ~2.00 & ~2915   & 1457 \\
    \textbf{Image Edit Instruction} && 1959	& ~2.14 & ~4987   & 2335 \\
    \textbf{Visual Story Telling}   && ~400	& ~4.50 & ~4338   & ~964 \\
    \textbf{Visual Cloze}           && ~200	& ~4.50 & ~8694   & 1932 \\
    \textbf{Text-rich VQA}          && ~400	& ~2.26 & ~4988   & 2210 \\
    \textbf{Multi-image VQA}        && ~400	& ~3.51 & ~5167   & 1472 \\
    \textbf{Puzzle}                 && 1400	& \textbf{12.00} & ~3072   & ~256 \\
    \textbf{NLVR2}                  && 6967	& ~2.00 & ~4218   & 2109 \\
    \textbf{Q-Bench}                && 1000	& ~2.00 & ~4742   & 2371 \\
    \midrule
    \textbf{ComPairs}               && ~532	& ~6.00 & \textbf{14619}  & \textbf{2436} \\
    \bottomrule
    \end{tabular}
    }
    \caption{Comparison in terms of visual input dimensions in popular multi-image datasets, gathered in the LLaVA-Interleave Benchmark, and in our proposed ComPairs dataset.}
    \label{tab:datasets_comparison}
\end{table}

Starting from these lists of characteristics-based comparative descriptions, we obtain the samples for ComPairs, each relative to one of the characteristics of the products pairs. In total, the dataset contains 532 comparative questions concerning a total of 272 unique characteristics of 88 different product types.

The multi-image sets comparison task is formulated as a visual-textual prompt in the form ``\textit{Given the images of Product 1:\token{img_1} and the images of Product 2:\token{img_2}, can you compare them in terms of \token{characteristic}?}'', where \token{img_1} and \token{img_2} are the images relative to Product 1 and Product 2, respectively. The LMM tasked with this prompt is expected to generate a short comparison of the two image sets in terms of the specified characteristic. 
One of the main characteristics of the ComPairs benchmark is that the images in the sets are high-resolution. 
In the context of LMMs, such images are often handled as multiple non-overlapping patches and a thumbnail (\ie, a lower-resolution version of the original image), as in the case of models that operate on image tiles, as for example InternVL~\cite{chen2024internvl}.  
Moreover, the product's images have a high variability in terms of density of information contained. Specifically, there are simple images of products (\eg, a chair, a soap dispenser, a pair of socks) where the high resolution could be harmful to the model due to the overload of information, and more complex images (\eg, the close-up of the back of a book, or the nutritional table of a chocolate bar) where the high resolution is necessary to extract the needed information.
These characterisitcs make ComPairs particularly challenging for LMMs and stress the need for a flexible visual token reduction strategy. 

To provide a data-driven motivation of the challenges posed by ComPairs in terms of input to be handled, we report in~\Cref{tab:datasets_comparison} a comparison between our proposed dataset and multi-modal, multi-image datasets belonging to the LLaVA-Interleave Benchmark~\cite{li2024llavaNI}, which is established when evaluating LMMs. 
As we can see, a sample in ComPairs entails handling $6$ images at a time, which is more than the average number of images involved on most of the datasets collected in LLaVA-Interleave Benchmark. 
More importantly, the images involved in each sample are represented with a number of tokens which is one order of magnitude bigger than the average number of tokens needed in the other tasks. 
For this reason, we argue that ComPairs represents a scenario in which visual token reduction is particularly interesting to explore.
\section{Experiments}
\begin{table}[t]
    \centering
    \resizebox{\columnwidth}{!}{%
    \begin{tabular}{l l l}
    \toprule
     & \textbf{Visual Encoder} & \textbf{LLM}\\
     \midrule
        \textbf{InternVL2‑4B}  & InternViT 300M 448px	    & Phi3 Mini 128k Instruct    \\
        \textbf{InternVL2‑8B}  & InternViT 300M 448px	    & InternLM2.5 7B Chat         \\
        \textbf{InternVL2‑26B} & InternViT 6B 448px V1.5    & InternLM2 Chat 20B          \\
        \textbf{InternVL2-76B} & InternViT 6B 448px V1.5	& Hermes2 Theta Llama 3 70B  \\
        \textbf{LLaVA-Next-Interleave} &  SigLIP 400M 384px   & Qwen1.5 7B \\
        \textbf{Phi3.5-Vision-Instruct} &  CLIP ViT-L/14	& Phi-3 Mini-128K-Instruct  \\
    \bottomrule
    \end{tabular}
    }
    \caption{Visual Encoder and LLM backbones in the considered LLMs.}
    \label{tab:llms}
\end{table}

We validate the effectiveness of~\method on the multi-modal multi-image datasets collected in the LLava-Interleave Benchmark~\cite{li2024llavaNI} and the ComPairs Benchmark.

\tit{LMM Architectures}
Given their capacity to handle high-dimensional visual and textual inputs (like multiple multi-patch images), we consider the models of the \emph{InternVL2 family}~\cite{chen2024internvl}. This is a series of models of varying sizes to balance computational efficiency and performance. 
Specifically, we consider both the larger variants (InternVL2-76B and InternVL2-26B) and the more compact ones (InternVL2-8B and InternVL2-4B), which are suitable for scenarios of computational resources availability. 
Moreover, we consider \emph{Phi-3.5-Vision}~\cite{abdin2024phi}, which also handles high-resolution images presented in a multi-patch format, and \emph{LLaVA-Next-Interleave}~\cite{li2024llavaNI}, which is trained on a massive dataset of diverse multi-image, multi-frame, multi-patch, multi-view multi-modal tasks, unified in a visual-textual interleaved data format. 
The specific Visual Encoder and LLM featured in the considered LMMs are reported in~\Cref{tab:llms}.
Note that our work evaluates on a diverse set of model architectures from different families, in contrast with previous work~\cite{arif2024hired, shang2024llavaPM} which focused on a single model family.

\begin{table*}[]
    \centering
    \resizebox{\textwidth}{!}{%
    \begin{tabular}{l c cc c cc c cc c cc}
    \toprule
    & 
    & \multicolumn{2}{c}{\textbf{InternVL2-4B}} &
    & \multicolumn{2}{c}{\textbf{InternVL2-8B}} &
    & \multicolumn{2}{c}{\textbf{InternVL2-26B}} &
    & \multicolumn{2}{c}{\textbf{InternVL2-76B}} \\
    \cmidrule{3-4} \cmidrule{6-7} \cmidrule{9-10} \cmidrule{12-13}
    & 
    & \textbf{w/o \method} & \textbf{w/ \method} &
    & \textbf{w/o \method} & \textbf{w/ \method} &
    & \textbf{w/o \method} & \textbf{w/ \method} &
    & \textbf{w/o \method} & \textbf{w/ \method} \\
    \midrule
    \textbf{Spot the Difference}    && 17.06 & 16.79 && 18.81 & 17.59 && 17.93 & 18.93 && 17.45 & 15.88 \\
    \textbf{Image Edit Instruction} && 10.47 & ~9.78 && 10.58 & 10.54 && 11.19 & 10.79 && 11.73 & 11.55 \\
    \textbf{Visual Story Telling}   && 23.32 & 23.99 && 19.70 & 19.42 && 25.98 & 26.57 && 21.44 & 23.72 \\
    \textbf{Visual Cloze}           && 19.00 & 29.50 && 25.00 & 20.50 && 19.50 & 19.00 && 14.00 & 13.50 \\
    \textbf{Text-rich VQA}          && 42.41 & 42.43 && 46.07 & 44.50 && 53.12 & 41.92 && 44.51 & 41.66 \\
    \textbf{Multi-image VQA}        && 46.25 & 52.25 && 29.50 & 35.50 && 12.75 & 17.50 && 21.00 & 25.75 \\
    \textbf{Puzzle}                 && 25.07 & 21.14 && 19.71 & 21.57 && 21.36 & 21.50 && 23.14 & 23.64 \\
    \textbf{NLVR2}                  && 64.03 & 66.50 && 17.73 & 19.39 && 20.86 & 20.12 && 78.51 & 73.37 \\
    \textbf{Q-Bench}                && 56.50 & 55.40 && 35.80 & 33.50 && 41.90 & 42.70 && 60.00 & 55.70 \\
    \textbf{\textit{Overall}}       && 33.79 & 35.31 && 24.77 & 24.72 && 24.95 & 24.34 && 32.42 & 31.64 \\
    \midrule
    \textbf{Avg used tokens [\%]}   && 100.00 & 41.03 && 100.00 & 40.62 && 100.00 & 42.38 && 100.00 & 35.19 \\
    \textbf{Std used tokens [\%]}   &&  0.00 & 28.18 &&  0.00 & 31.64 &&  0.00 & 32.79 &&  0.00 & 34.43 \\
    \textbf{Max GPU memory [GB]}    && 30.46 & 10.35 && 32.42 & 19.00 && 80.91 & 58.40 && 184.98 & 152.97 \\
    \textbf{Runtime [s]}            &&  3.91 &  4.01 &&  3.89 &  4.56 &&  7.69 &  7.59 && 21.63 & 15.21 \\
    \bottomrule
    \end{tabular}
    }
    \caption{Effect of the LMM sizes both in terms of accuracy (in percentage) and efficiency on the LLaVA-Interleave Benchmark. We used $1, 2, 4, 8$ NVIDIA L40S GPUs with 48GB of memory for the four models, respectively.}
    \label{tab:backbone_sizes}
\end{table*}

\tit{Performance Evaluation}
We express the performance both in terms of efficiency gain and accuracy. In particular, we measure the maximum GPU memory consumption of different LMMs combined with our \method method and alternative strategies for visual token reduction. Moreover, we consider the runtime of such models when performing the considered tasks on NVIDIA L40S GPUs. As for the accuracy computation, we observe that oftentimes the LLM inside the LMMs generates an answer that is semantically correct but is not necessarily verbatim identical to the ground truth answer. Therefore, simply comparing the reference and predicted answer word-by-word can lead to misleading results. For this reason, we compute the accuracy of the considered models by tasking an external LLM (specifically, Llama 3 8B Instruct) to compare their given answers to the ground truth. Then, we express the accuracy as the percentage of answers that the LLM has labeled as corresponding to the ground truth answer.

\subsection{Results on the LLaVA-Interleave Benchmark}
We consider the multi-image datasets collected in the LLaVA-Interleave Benchmark. These are relative to different multi-modal tasks performed on different numbers of images, as reported in~\Cref{tab:datasets_comparison}. 

\tit{\method on Different-sized LMMs}
The LMMs in the InternVL~\cite{chen2024internvl} family consume a large number of visual tokens, have been implemented with similar architectural and training choices, and come in different sizes, depending on the Visual Encoder and LLM backbone used (see~\Cref{tab:llms}). 
For this reason, we consider these models for analyzing the effect of \method when applied to increasingly large models and report the results of this study in~\Cref{tab:backbone_sizes}. As we can see, applying our token reduction strategy does not significantly affect the performance across model sizes, ranging from an average improvement of +1.52\% for the smallest considered model (InternVL2‑4B) to a limited reduction of -0.78\% for the largest model (InternVL2-76B). 
Most importantly, \method consistently reduces the number of used tokens by $\sim$60\%, thus reducing the maximum GPU memory required for factors ranging from 17.30\% for InternVL2-76B to 66.22\% for the already compact InternVL2‑4B, further reducing the computational requirements of this model and thus, enhancing its applicability in low-resource scenarios. As for the runtime, applying \method inevitably introduces some overhead. However, such overhead is negligible \wrt~the overall computation time of the smaller models (it is at most +0.67s, on InternVL2‑26B). Interestingly, the most relevant effect on InternVL2-76B is the runtime reduction obtained with \method (-6.42 s). Arguably, this is because this variant features the largest LLM in the InternVL family, and thus, reducing the number of visual tokens it has to consider noticeably speeds up the computation.

\tit{Comparison with Other Early Dropping Strategies}
From~\Cref{tab:backbone_sizes}, it also emerges that the best-performing model on average is InternVL2-4B, further boosted by applying our \method algorithm. For this reason, in the rest of our analysis on the LLaVA-Interleave Benchmark, we consider this LMM. Here, we combine it with different visual token reduction strategies to assess the effectiveness of \method with respect to alternatives (see~\Cref{tab:best_reduction}). Specifically, we consider a \emph{Random Sampling} as a baseline entailing token reduction via unprincipled selection. Moreover, we consider the recently-proposed \emph{HiRED}~\cite{arif2024hired} token selection approach, which is a state-of-the-art method for dropping redundant tokens. 
Both for Random sampling and HiRED the desired number of tokens to be fed to the LLM (\ie~a budget) has to be explicitly specified as input. 
On the other hand, \method automatically determines the final number of tokens based on the threshold $\tau$ (which depends on the number of input tokens). Here, we assign the Random sampling a budget that makes them retain a number of tokens that is comparable to the reduction operated by \method, while for HiRED we use the best value for the budget as indicated by its authors. 
The result of this comparison can be observed in~\Cref{tab:best_reduction}. From the Table it emerges that the specific strategy adopted for reducing the number of the visual tokens has an effect on the overall performance. Specifically, simply dropping tokens (either randomly or based on their CLS-attention) can lead to a performance drop, even if contained. On the other hand, maintaining most of the tokens information via fusion, as in~\method, has a positive effect on the performance. This is especially true for complex tasks such as visual sequence completion (Visual Cloze) and visual question answering (Text-rich VQA and Multi-image VQA), involving a large number of visual tokens and good reasoning capabilities. 
As for the saved GPU memory, it directly depends on the number of visual tokens used. Therefore, HiRED leads to the lowest memory consumption with the budget hyperparameter recommended by its authors (-74.62\% of the total memory used by the vanilla InternVL2-4B, versus -66.78\%  and -66.02\% obtained with Random sampling and \method) at the cost of the most noticeable performance drop. Moreover, \method runtime (4.01s) is comparable to the vanilla model (3.91s). Note that HiRED and Random sampling are $sim$1.5s faster than \method. Arguably, this is due to the fact that these two approaches simply drop visual tokens, allowing them to reduce the runtime but causing some performance drop, which does not happen with~\method.

\begin{table}[t]
    \centering
    \resizebox{\columnwidth}{!}{%
    \begin{tabular}{l c cccc}
    \toprule
    & 
    & \makecell{\textbf{All tokens}} 
    & \makecell{\textbf{Random}\\ \textbf{sampling}}
    & \makecell{\textbf{HiRED}}
    & \makecell{\textbf{\method}}\\
    \midrule
    \textbf{Spot the Difference}    & & 17.06 & 15.59 & 16.74 & 16.79 \\
    \textbf{Image Edit Instruction} & & 10.47 & 10.15 & ~9.91 &  9.78 \\
    \textbf{Visual Story Telling}   & & 23.32 & 24.05 & 24.01 & 23.99 \\
    \textbf{Visual Cloze}           & & 19.00 & 21.00 & 20.00 & 29.50 \\
    \textbf{Text-rich VQA}          & & 42.41 & 39.47 & 37.70 & 42.43 \\
    \textbf{Multi-image VQA}        & & 46.25 & 48.75 & 45.25 & 52.25 \\
    \textbf{Puzzle}                 & & 25.07 & 20.07 & 20.71 & 21.14 \\
    \textbf{NLVR2}                  & & 64.03 & 68.65 & 65.14 & 66.50 \\
    \textbf{Q-Bench}                & & 54.50 & 55.40 & 55.80 & 55.40 \\
    \textbf{\textit{Overall}}       & & 33.79 & 33.68 & 32.81 & 35.31 \\
    \midrule
    \textbf{Avg used tokens [\%]}   & &100.00 & 40.04 & 20.00 & 41.03 \\
    \textbf{Std used tokens [\%]}   & &~~0.00 & ~2.34 & ~0.00 & 28.18 \\
    \textbf{Max GPU memory [GB]}    & &~30.46 & 10.12 & ~7.73 & 10.35 \\
    \textbf{Runtime [s]}            & &~~3.91 & ~2.72 & ~2.55 & ~4.01 \\
    \bottomrule
    \end{tabular}
    }
    \caption{Best reduction strategy with fixed LLM (InternVL2-4B) on the LLaVA-Interleave Benchmark. }
    \label{tab:best_reduction}
\end{table}

\begin{figure}[t]
    \centering
    \includegraphics[width=\linewidth]{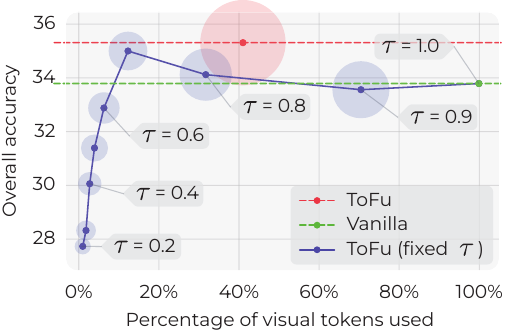}
    \caption{Overall accuracy (y-axis) and percentage of tokens (x-axis) used by InternVL2-4B on the LLava-Interleave Benchmark. We compare our \method, which has a dynamic threshold (red line), against a variant of \method with fixed $\tau$ (blue line) that applies the same threshold $\tau$ for each image in the benchmark and the vanilla InternVL2-4B which uses all tokens (green line).
    For each value of the threshold, we indicate the corresponding average number of tokens and the standard deviation as a semi-transparent circle.}
    \label{fig:lib_sweep}
\end{figure}
    
\tit{Effect of the Dynamic Threshold Selection}
The \method algorithm relies on a threshold $\tau$. Instead of assigning a fixed value regardless of the input length, we automatically adjust it based on the number of visual tokens the algorithm has to deal with, as described in~\Cref{sec:tofu}. 
To validate the effectiveness of this strategy, we perform an ablation in which we apply~\method to InternVL2‑4B with a threshold value that stays the same for all the tasks (and thus, visual input length ) of the LaVA-Interleave Benchmark. We run multiple instances of this combination, by varying the value of the threshold at each run. In~\Cref{fig:lib_sweep}, we report the result of this ablation in terms of average accuracy with respect to the percentage number of used tokens. For reference, we also report the performance of the vanilla InternVL2‑4B, using all the tokens, and that of InternVL2‑4B combined with our complete~\method method, entailing dynamic threshold setting. As we can see, adapting the value of the threshold leads to a varied token reduction of 41.03\%$\pm$28.18\% and to consistently superior performance both with respect to the vanilla InternVL2‑4B and to the non-adaptive threshold strategy.

\begin{table}
    \centering
    \setlength{\tabcolsep}{.18em}
    \resizebox{\columnwidth}{!}{%
    \begin{tabular}{l c  cc  c cc c cc}
    \toprule
    & 
    & \multicolumn{2}{c}{\makecell{\textbf{All tokens}}} &
    & \multicolumn{2}{c}{\makecell{\textbf{Random} \\ \textbf{sampling}}} &
    & \multicolumn{2}{c}{\makecell{\textbf{\method}}}\\
    \cmidrule{3-4} \cmidrule{6-7} \cmidrule{9-10}
    & 
    & Acc. & \% tokens &
    & Acc. & \% tokens &
    & Acc. & \% tokens \\
    \midrule
    \textbf{LLaVa-Next-Interleave} & &  3.95 &100.00 & &  4.70 & 40.00 & &  5.45 & 39.59 \\
    \textbf{InternVL2-76B}         & & 18.98 &100.00 & & 16.73 & 39.86 & & 17.29 & 48.83 \\
    \textbf{InternVL2‑26B}         & & 13.16 &100.00 & & 16.54 & 39.95 & & 14.66 & 51.71 \\
    \textbf{InternVL2‑8B}          & &  9.02 &100.00 & & 11.09 & 50.11 & & 23.68 & 43.17 \\
    \textbf{InternVL2‑4B}          & & 12.03 &100.00 & &  9.77 & 39.85 & & 14.85 & 43.11 \\
    \textbf{Phi3.5-Vision-Instruct}& & 10.53 &100.00 & &  8.65 & 40.02 & & 11.84 & 35.30 \\
    \bottomrule
    \end{tabular}
    }
    \caption{Best reduction strategy on ComPairs with different LMMs. For a fair comparison with random pruning, we prune in order to keep a number of tokens similar to that used in~\method.}
    \label{tab:compairs}
\end{table}

\begin{figure*}[t]
    \centering
    \includegraphics[width=.97\linewidth]{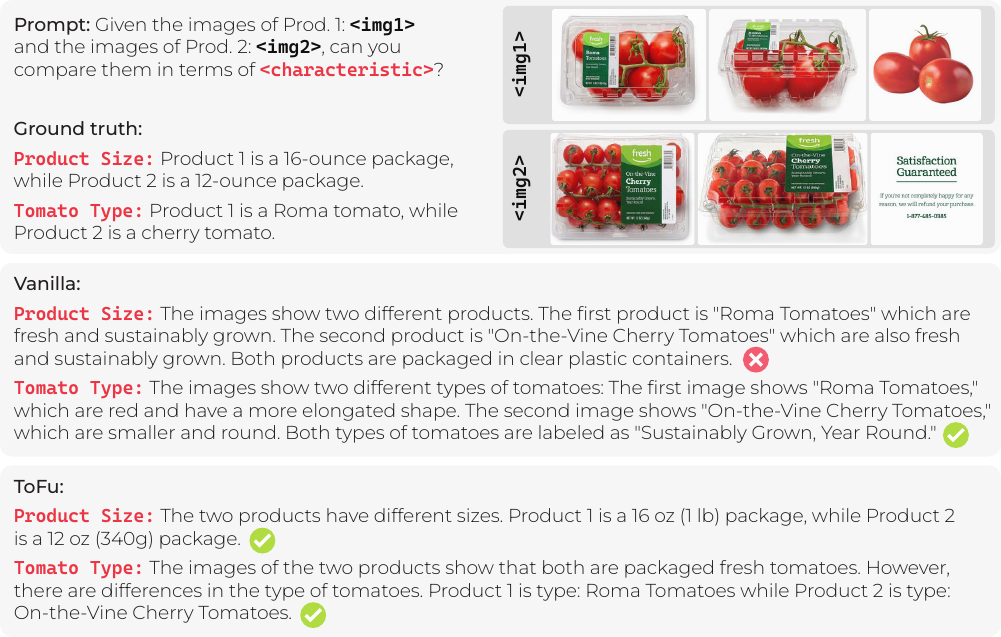}
    \caption{Sample from the ComPairs dataset where we show the benefits and the results obtained with the \method algorithm with respect to the vanilla approach. For space reasons, we have abbreviated the word ``Product'' in the image to ``Prod.''}
    \label{fig:sample}
\end{figure*}

\subsection{Results on the ComPairs Benchmark}
The ComPairs dataset serves as a stress-test for our token reduction strategy, since it entails dealing with a much larger number of visual tokens compared to existing benchmarks. 
\Cref{tab:compairs} shows results of different LMMs, both as-is and combined with the \method algorithm and random sampling while keeping comparable the number of retained visual tokens. 
We considered random sampling as a competitor here instead of HiRED because~\Cref{tab:best_reduction} shows that it is the second-best method overall. Moreover, it is easily applicable to any architecture without any dependency on other components of the LMM, unlike HiRED, which relies on the self-attention maps of the visual encoder.
Nonetheless, we have run an experiment using HiRED with InternVL2-4B and obtained 9.59\% Accuracy which is similar to random sampling and lower than~\method as shown in~\Cref{tab:compairs} (c.f., row 5). 
The complete results of this analysis are reported in~\Cref{tab:compairs}. 
As it can be observed, applying \method leads to performance improvement, for almost all the considered LMMs (with the best-performing combination being with IntarnVL2-8B, 23.68\% accuracy). 
It also emerges that for the ComPair task, despite being characterized by a large number of likely redundant tokens, simply dropping some tokens is not an effective strategy. In fact, random sampling leads to a performance drop for almost all the considered LMMs. 
This analysis confirms that ComPairs is a suitable testbench for LMMs aimed at handling multi-patch, multi-image tasks and that it helps highlight the importance of applying visual token reduction strategies in such high-dimensional inputs. Moreover, these results demonstrate the effectiveness of our proposed \method algorithm. 
Finally, we show some qualitative results on two samples of the ComPAirs dataset, obtained from the same sets of images in~\Cref{fig:sample}. We compare InternVL2-4B vanilla (\ie, without any toke reduction strategy) and InternVL2-4B with our \method algorithm. It can be observed that the combination with \method increases the comparison performance, especially regarding characteristics that require fine-grained visual information (such as those that can be determined from writings and labels). Further qualitative examples can be found in the Appendix.

\section{Conclusion}
In this work, we have proposed \method, a training-free, visual encoder-agnostic approach to enhance the efficiency of LMMs. Specifically, the proposed algorithm fuses together similar visual tokens embeddings before feeding them to the LLM, thus reducing its prefix context length without significant information loss. In this way, we are able to reduce the GPU memory requirements of LMMs while maintaining the performance with respect to using all the tokens. Extensive evaluation on several multi-image datasets and on a newly-proposed challenging dataset, dubbed ComPairs, demonstrate the effectiveness of our approach over a diverse set of LMM architectures.

\section*{Acknowledgement}
This work was supported by the ELSA - European Lighthouse on Secure and
Safe AI project funded by the EU (GA 101070617) and the PNRR project Italian Strengthening of Esfri RI Resilience (ITSERR) funded by the European Union – NextGenerationEU (CUP: B53C22001770006).

{
    \small
    \bibliographystyle{ieeenat_fullname}
    \bibliography{main}

\begin{thebibliography}{45}
\providecommand{\natexlab}[1]{#1}
\providecommand{\url}[1]{\texttt{#1}}
\expandafter\ifx\csname urlstyle\endcsname\relax
  \providecommand{\doi}[1]{doi: #1}\else
  \providecommand{\doi}{doi: \begingroup \urlstyle{rm}\Url}\fi

\bibitem[Abdin et~al.(2024)Abdin, Aneja, Awadalla, Awadallah, Awan, Bach, Bahree, Bakhtiari, Bao, Behl, et~al.]{abdin2024phi}
Marah Abdin, Jyoti Aneja, Hany Awadalla, Ahmed Awadallah, Ammar~Ahmad Awan, Nguyen Bach, Amit Bahree, Arash Bakhtiari, Jianmin Bao, Harkirat Behl, et~al.
\newblock Phi-3 technical report: A highly capable language model locally on your phone.
\newblock \emph{arXiv preprint arXiv:2404.14219}, 2024.

\bibitem[Agrawal et~al.(2024)Agrawal, Antoniak, Hanna, Chaplot, Chudnovsky, Garg, Gervet, Ghosh, H{\'e}liou, Jacob, et~al.]{agrawal2024pixtral}
Pravesh Agrawal, Szymon Antoniak, Emma~Bou Hanna, Devendra Chaplot, Jessica Chudnovsky, Saurabh Garg, Theophile Gervet, Soham Ghosh, Am{\'e}lie H{\'e}liou, Paul Jacob, et~al.
\newblock Pixtral 12b.
\newblock \emph{arXiv preprint arXiv:2410.07073}, 2024.

\bibitem[Anthropic(2024)]{anthropic2024claude}
AI Anthropic.
\newblock The claude 3 model family: Opus, sonnet, haiku.
\newblock \emph{Claude-3 Model Card}, 2024.

\bibitem[Arif et~al.(2024)Arif, Yoon, Nikolopoulos, Vandierendonck, John, and Ji]{arif2024hired}
Kazi Hasan~Ibn Arif, JinYi Yoon, Dimitrios~S Nikolopoulos, Hans Vandierendonck, Deepu John, and Bo Ji.
\newblock {HiRED: Attention-Guided Token Dropping for Efficient Inference of High-Resolution Vision-Language Models in Resource-Constrained Environments}.
\newblock \emph{arXiv preprint arXiv:2408.10945}, 2024.

\bibitem[Bansal et~al.(2020)Bansal, Zhang, and Chellappa]{bansal2020visual}
Ankan Bansal, Yuting Zhang, and Rama Chellappa.
\newblock Visual question answering on image sets.
\newblock In \emph{ECCV}. Springer, 2020.

\bibitem[Bolya et~al.(2023)Bolya, Fu, Dai, Zhang, Feichtenhofer, and Hoffman]{bolya2023token}
Daniel Bolya, Cheng-Yang Fu, Xiaoliang Dai, Peizhao Zhang, Christoph Feichtenhofer, and Judy Hoffman.
\newblock {Token Merging: Your ViT But Faster}.
\newblock In \emph{ICLR}, 2023.

\bibitem[Buch et~al.(2022)Buch, Eyzaguirre, Gaidon, Wu, Fei-Fei, and Niebles]{buch2022revisiting}
Shyamal Buch, Crist{\'o}bal Eyzaguirre, Adrien Gaidon, Jiajun Wu, Li Fei-Fei, and Juan~Carlos Niebles.
\newblock Revisiting the "video" in video-language understanding.
\newblock In \emph{CVPR}, 2022.

\bibitem[Cai et~al.(2024)Cai, Yang, Gao, and Lee]{cai2024matryoshka}
Mu Cai, Jianwei Yang, Jianfeng Gao, and Yong~Jae Lee.
\newblock {Matryoshka Multimodal Models}.
\newblock \emph{arXiv preprint arXiv:2405.17430}, 2024.

\bibitem[Chen et~al.(2024{\natexlab{a}})Chen, Zhao, Liu, Bai, Lin, Zhou, and Chang]{chen2024image}
Liang Chen, Haozhe Zhao, Tianyu Liu, Shuai Bai, Junyang Lin, Chang Zhou, and Baobao Chang.
\newblock {An Image is Worth 1/2 Tokens After Layer 2: Plug-and-Play Inference Acceleration for Large Vision-Language Models}.
\newblock \emph{arXiv preprint arXiv:2403.06764}, 2024{\natexlab{a}}.

\bibitem[Chen et~al.(2024{\natexlab{b}})Chen, Wang, Tian, Ye, Gao, Cui, Tong, Hu, Luo, Ma, Ma, Wang, Dong, Yan, Guo, He, Shi, Jin, Xu, Wang, Wei, Li, Zhang, Zhang, Cai, Wen, Yan, Dou, Lu, Zhu, Lu, Lin, Qiao, Dai, and Wang]{chen2024far}
Zhe Chen, Weiyun Wang, Hao Tian, Shenglong Ye, Zhangwei Gao, Erfei Cui, Wenwen Tong, Kongzhi Hu, Jiapeng Luo, Zheng Ma, Ji Ma, Jiaqi Wang, Xiaoyi Dong, Hang Yan, Hewei Guo, Conghui He, Botian Shi, Zhenjiang Jin, Chao Xu, Bin Wang, Xingjian Wei, Wei Li, Wenjian Zhang, Bo Zhang, Pinlong Cai, Licheng Wen, Xiangchao Yan, Min Dou, Lewei Lu, Xizhou Zhu, Tong Lu, Dahua Lin, Yu Qiao, Jifeng Dai, and Wenhai Wang.
\newblock How far are we to gpt-4v? closing the gap to commercial multimodal models with open-source suites, 2024{\natexlab{b}}.

\bibitem[Chen et~al.(2024{\natexlab{c}})Chen, Wu, Wang, Su, Chen, Xing, Zhong, Zhang, Zhu, Lu, et~al.]{chen2024internvl}
Zhe Chen, Jiannan Wu, Wenhai Wang, Weijie Su, Guo Chen, Sen Xing, Muyan Zhong, Qinglong Zhang, Xizhou Zhu, Lewei Lu, et~al.
\newblock Internvl: Scaling up vision foundation models and aligning for generic visual-linguistic tasks.
\newblock In \emph{Proceedings of the IEEE/CVF Conference on Computer Vision and Pattern Recognition}, pages 24185--24198, 2024{\natexlab{c}}.

\bibitem[Chu et~al.(2023{\natexlab{a}})Chu, Qiao, Lin, Xu, Yang, Hu, Wei, Zhang, Zhang, Wei, and Shen]{chu2023mobilevlmfaststrong}
Xiangxiang Chu, Limeng Qiao, Xinyang Lin, Shuang Xu, Yang Yang, Yiming Hu, Fei Wei, Xinyu Zhang, Bo Zhang, Xiaolin Wei, and Chunhua Shen.
\newblock Mobilevlm : A fast, strong and open vision language assistant for mobile devices, 2023{\natexlab{a}}.

\bibitem[Chu et~al.(2023{\natexlab{b}})Chu, Qiao, Lin, Xu, Yang, Hu, Wei, Zhang, Zhang, Wei, et~al.]{chu2023mobilevlm}
Xiangxiang Chu, Limeng Qiao, Xinyang Lin, Shuang Xu, Yang Yang, Yiming Hu, Fei Wei, Xinyu Zhang, Bo Zhang, Xiaolin Wei, et~al.
\newblock {MobileVLM: A Fast, Reproducible and Strong Vision Language Assistant for Mobile Devices}.
\newblock \emph{arXiv preprint arXiv:2312.16886}, 2023{\natexlab{b}}.

\bibitem[Collins et~al.(2022)Collins, Goel, Deng, Luthra, Xu, Gundogdu, Zhang, Yago~Vicente, Dideriksen, Arora, Guillaumin, and Malik]{collins2022abo}
Jasmine Collins, Shubham Goel, Kenan Deng, Achleshwar Luthra, Leon Xu, Erhan Gundogdu, Xi Zhang, Tomas~F Yago~Vicente, Thomas Dideriksen, Himanshu Arora, Matthieu Guillaumin, and Jitendra Malik.
\newblock Abo: Dataset and benchmarks for real-world 3d object understanding.
\newblock \emph{CVPR}, 2022.

\bibitem[Dubey et~al.(2024)Dubey, Jauhri, Pandey, Kadian, Al-Dahle, Letman, Mathur, Schelten, Yang, Fan, et~al.]{dubey2024llama}
Abhimanyu Dubey, Abhinav Jauhri, Abhinav Pandey, Abhishek Kadian, Ahmad Al-Dahle, Aiesha Letman, Akhil Mathur, Alan Schelten, Amy Yang, Angela Fan, et~al.
\newblock The llama 3 herd of models.
\newblock \emph{arXiv preprint arXiv:2407.21783}, 2024.

\bibitem[Fu et~al.(2024)Fu, Chen, Shen, Qin, Zhang, Lin, Yang, Zheng, Li, Sun, Wu, and Ji]{fu2024mmecomprehensiveevaluationbenchmark}
Chaoyou Fu, Peixian Chen, Yunhang Shen, Yulei Qin, Mengdan Zhang, Xu Lin, Jinrui Yang, Xiawu Zheng, Ke Li, Xing Sun, Yunsheng Wu, and Rongrong Ji.
\newblock Mme: A comprehensive evaluation benchmark for multimodal large language models, 2024.

\bibitem[Hong et~al.(2023)Hong, Lin, Du, Chen, Tenenbaum, and Gan]{hong20233d}
Yining Hong, Chunru Lin, Yilun Du, Zhenfang Chen, Joshua~B Tenenbaum, and Chuang Gan.
\newblock 3d concept learning and reasoning from multi-view images.
\newblock In \emph{CVPR}, 2023.

\bibitem[Jiang et~al.(2024)Jiang, He, Zeng, Wei, Ku, Liu, and Chen]{jiang2024mantis}
Dongfu Jiang, Xuan He, Huaye Zeng, Cong Wei, Max Ku, Qian Liu, and Wenhu Chen.
\newblock Mantis: Interleaved multi-image instruction tuning, 2024.

\bibitem[Kitaev et~al.(2019)Kitaev, Kaiser, and Levskaya]{kitaev2019reformer}
Nikita Kitaev, Lukasz Kaiser, and Anselm Levskaya.
\newblock {Reformer: The Efficient Transformer}.
\newblock In \emph{ICLR}, 2019.

\bibitem[Li et~al.(2024{\natexlab{a}})Li, Ge, Ge, Wang, Wang, Zhang, and Shan]{Li_2024_CVPR}
Bohao Li, Yuying Ge, Yixiao Ge, Guangzhi Wang, Rui Wang, Ruimao Zhang, and Ying Shan.
\newblock Seed-bench: Benchmarking multimodal large language models.
\newblock In \emph{Proceedings of the IEEE/CVF Conference on Computer Vision and Pattern Recognition (CVPR)}, pages 13299--13308, 2024{\natexlab{a}}.

\bibitem[Li et~al.(2024{\natexlab{b}})Li, Zhang, Zhang, Zhang, Li, Li, Ma, and Li]{li2024llavaNI}
Feng Li, Renrui Zhang, Hao Zhang, Yuanhan Zhang, Bo Li, Wei Li, Zejun Ma, and Chunyuan Li.
\newblock {LLaVA-NeXT-Interleave: Tackling Multi-image, Video, and 3D in Large Multimodal Models}.
\newblock \emph{arXiv preprint arXiv:2407.07895}, 2024{\natexlab{b}}.

\bibitem[Li et~al.(2024{\natexlab{c}})Li, Chen, Cai, Chen, Hong, Chen, Shen, and Gan]{li2024flexattention}
Junyan Li, Delin Chen, Tianle Cai, Peihao Chen, Yining Hong, Zhenfang Chen, Yikang Shen, and Chuang Gan.
\newblock {FlexAttention for Efficient High-Resolution Vision-Language Models}.
\newblock \emph{arXiv preprint arXiv:2407.20228}, 2024{\natexlab{c}}.

\bibitem[Li et~al.(2023)Li, He, Wang, Li, Wang, Luo, Wang, Wang, and Qiao]{li2023videochat}
KunChang Li, Yinan He, Yi Wang, Yizhuo Li, Wenhai Wang, Ping Luo, Yali Wang, Limin Wang, and Yu Qiao.
\newblock Videochat: Chat-centric video understanding.
\newblock \emph{arXiv preprint arXiv:2305.06355}, 2023.

\bibitem[Liu et~al.(2023{\natexlab{a}})Liu, Li, Li, and Lee]{liu2023improvedllava}
Haotian Liu, Chunyuan Li, Yuheng Li, and Yong~Jae Lee.
\newblock Improved baselines with visual instruction tuning, 2023{\natexlab{a}}.

\bibitem[Liu et~al.(2023{\natexlab{b}})Liu, Li, Wu, and Lee]{liu2023llava}
Haotian Liu, Chunyuan Li, Qingyang Wu, and Yong~Jae Lee.
\newblock Visual instruction tuning, 2023{\natexlab{b}}.

\bibitem[Liu et~al.(2023{\natexlab{c}})Liu, Wu, and Guo]{liu2023adaptive}
Xiangcheng Liu, Tianyi Wu, and Guodong Guo.
\newblock {Adaptive Sparse ViT: Towards Learnable Adaptive Token Pruning by Fully Exploiting Self-Attention}.
\newblock In \emph{IJCAI}, pages 1222--1230, 2023{\natexlab{c}}.

\bibitem[Liu et~al.(2024)Liu, Duan, Zhang, Li, Zhang, Zhao, Yuan, Wang, He, Liu, et~al.]{liu2025mmbench}
Yuan Liu, Haodong Duan, Yuanhan Zhang, Bo Li, Songyang Zhang, Wangbo Zhao, Yike Yuan, Jiaqi Wang, Conghui He, Ziwei Liu, et~al.
\newblock Mmbench: Is your multi-modal model an all-around player?
\newblock In \emph{European Conference on Computer Vision}. Springer, 2024.

\bibitem[Ma et~al.(2024)Ma, Wang, Ma, Wang, Wang, Huang, Dong, Wang, Xue, and Wei]{ma2024era}
Shuming Ma, Hongyu Wang, Lingxiao Ma, Lei Wang, Wenhui Wang, Shaohan Huang, Li Dong, Ruiping Wang, Jilong Xue, and Furu Wei.
\newblock {The Era of 1-bit LLMs: All Large Language Models are in 1.58 Bits}.
\newblock \emph{arXiv e-prints}, pages arXiv--2402, 2024.

\bibitem[McKinzie et~al.(2024)McKinzie, Gan, Fauconnier, Dodge, Zhang, Dufter, Shah, Du, Peng, Weers, et~al.]{mckinzie2024mm1}
Brandon McKinzie, Zhe Gan, Jean-Philippe Fauconnier, Sam Dodge, Bowen Zhang, Philipp Dufter, Dhruti Shah, Xianzhi Du, Futang Peng, Floris Weers, et~al.
\newblock Mm1: Methods, analysis \& insights from multimodal llm pre-training.
\newblock \emph{arXiv preprint arXiv:2403.09611}, 2024.

\bibitem[OpenAI(2024)]{openai2024gpt4}
OpenAI.
\newblock Gpt-4 technical report, 2024.

\bibitem[Shang et~al.(2024)Shang, Cai, Xu, Lee, and Yan]{shang2024llavaPM}
Yuzhang Shang, Mu Cai, Bingxin Xu, Yong~Jae Lee, and Yan Yan.
\newblock {LLaVA-PruMerge: Adaptive Token Reduction for Efficient Large Multimodal Models}.
\newblock \emph{arXiv preprint arXiv:2403.15388}, 2024.

\bibitem[Tanaka et~al.(2023)Tanaka, Nishida, Nishida, Hasegawa, Saito, and Saito]{tanaka2023slidevqa}
Ryota Tanaka, Kyosuke Nishida, Kosuke Nishida, Taku Hasegawa, Itsumi Saito, and Kuniko Saito.
\newblock Slidevqa: A dataset for document visual question answering on multiple images.
\newblock In \emph{AAAI}, 2023.

\bibitem[Team(2024)]{geminiteam2024gemini}
Gemini Team.
\newblock Gemini 1.5: Unlocking multimodal understanding across millions of tokens of context, 2024.

\bibitem[Wang et~al.(2024{\natexlab{a}})Wang, Zhao, Liu, Chen, Zhuang, Gu, Guo, and Zhao]{wang2024large}
Maolin Wang, Yao Zhao, Jiajia Liu, Jingdong Chen, Chenyi Zhuang, Jinjie Gu, Ruocheng Guo, and Xiangyu Zhao.
\newblock {Large Multimodal Model Compression via Iterative Efficient Pruning and Distillation}.
\newblock 2024{\natexlab{a}}.

\bibitem[Wang et~al.(2024{\natexlab{b}})Wang, Bai, Tan, Wang, Fan, Bai, Chen, Liu, Wang, Ge, et~al.]{wang2024qwen2}
Peng Wang, Shuai Bai, Sinan Tan, Shijie Wang, Zhihao Fan, Jinze Bai, Keqin Chen, Xuejing Liu, Jialin Wang, Wenbin Ge, et~al.
\newblock Qwen2-vl: Enhancing vision-language model's perception of the world at any resolution.
\newblock \emph{arXiv preprint arXiv:2409.12191}, 2024{\natexlab{b}}.

\bibitem[Wang et~al.(2020)Wang, Li, Khabsa, Fang, and Ma]{wang2020linformer}
Sinong Wang, Belinda~Z Li, Madian Khabsa, Han Fang, and Hao Ma.
\newblock {Linformer: Self-Attention with Linear Complexity}.
\newblock \emph{arXiv preprint arXiv:2006.04768}, 2020.

\bibitem[Wu et~al.(2023)Wu, Fei, Qu, Ji, and Chua]{wu2023next}
Shengqiong Wu, Hao Fei, Leigang Qu, Wei Ji, and Tat-Seng Chua.
\newblock Next-gpt: Any-to-any multimodal llm.
\newblock \emph{arXiv preprint arXiv:2309.05519}, 2023.

\bibitem[Xie et~al.(2024)Xie, Zhang, Lin, Cao, and Ji]{xie2024advancing}
Jingjing Xie, Yuxin Zhang, Mingbao Lin, Liujuan Cao, and Rongrong Ji.
\newblock {Advancing Multimodal Large Language Models with Quantization-Aware Scale Learning for Efficient Adaptation}.
\newblock In \emph{ACM MM}, 2024.

\bibitem[Ye et~al.(2024)Ye, Xu, Liu, Hu, Yan, Qian, Zhang, Huang, and Zhou]{ye2024mplug}
Jiabo Ye, Haiyang Xu, Haowei Liu, Anwen Hu, Ming Yan, Qi Qian, Ji Zhang, Fei Huang, and Jingren Zhou.
\newblock mplug-owl3: Towards long image-sequence understanding in multi-modal large language models.
\newblock \emph{arXiv preprint arXiv:2408.04840}, 2024.

\bibitem[Yuan et~al.(2024)Yuan, Li, Huang, Ye, and Sun]{yuan2024tinygpt}
Zhengqing Yuan, Zhaoxu Li, Weiran Huang, Yanfang Ye, and Lichao Sun.
\newblock Tinygpt-v: Efficient multimodal large language model via small backbones.
\newblock In \emph{ICMLW}, 2024.

\bibitem[Zhang et~al.(2024{\natexlab{a}})Zhang, Zeng, Wang, and Lu]{zhang2024tinyllamaopensourcesmalllanguage}
Peiyuan Zhang, Guangtao Zeng, Tianduo Wang, and Wei Lu.
\newblock Tinyllama: An open-source small language model, 2024{\natexlab{a}}.

\bibitem[Zhang et~al.(2024{\natexlab{b}})Zhang, Lyu, Shao, Chen, Guan, and Nie]{zhang2024token}
Renshan Zhang, Yibo Lyu, Rui Shao, Gongwei Chen, Weili Guan, and Liqiang Nie.
\newblock {Token-level Correlation-guided Compression for Efficient Multimodal Document Understanding}.
\newblock \emph{arXiv preprint arXiv:2407.14439}, 2024{\natexlab{b}}.

\bibitem[Zhang et~al.()Zhang, Kishore, Wu, Weinberger, and Artzi]{zhangbertscore}
Tianyi Zhang, Varsha Kishore, Felix Wu, Kilian~Q Weinberger, and Yoav Artzi.
\newblock {BERTScore: Evaluating Text Generation with BERT}.
\newblock In \emph{ICLR}.

\bibitem[Zhou et~al.(2024)Zhou, Hu, Weng, Jia, Luo, Liu, Wu, and Huang]{zhou2024tinyllavaframeworksmallscalelarge}
Baichuan Zhou, Ying Hu, Xi Weng, Junlong Jia, Jie Luo, Xien Liu, Ji Wu, and Lei Huang.
\newblock Tinyllava: A framework of small-scale large multimodal models, 2024.

\bibitem[Zhu et~al.(2024)Zhu, Chen, Shen, Li, and Elhoseiny]{zhu2024minigpt}
Deyao Zhu, Jun Chen, Xiaoqian Shen, Xiang Li, and Mohamed Elhoseiny.
\newblock Mini{GPT}-4: Enhancing vision-language understanding with advanced large language models.
\newblock In \emph{ICLR}, 2024.

\end{thebibliography}
}



\appendix
\section{Further Examples from ComPairs}
\label{sec:further_compairs}
In this section, we report further examples and qualitative results from the ComPairs dataset. Specifically, we show pairs of image sets and ground truth answers relative to different characteristics of the objects in the sets. Recall that these characteristics and ground truth comparison answers have been obtained with an LLM (Llama3-8B) fed with the textual description associated with the products in the images. Moreover, we report the output of InternVL2-4B without any token reduction strategies (\textbf{InternVL2-4B Vanilla}) and of the same LMM combined with our visual token reduction approach (\textbf{InternVL2-4B + \method}) when prompted with the following prompt: "Given the images of Product 1: $\mathtt{{<}img1{>}}$ and the images of Product 2: $\mathtt{{<}img2{>}}$, can you compare them in terms of \textcolor{RoyalBlue}{$\mathtt{{<}characteristic{>}}$}?", where $\mathtt{{<}img1{>}}$ and $\mathtt{{<}img2{>}}$ are sequences of visual tokens obtained from the images in the two sets under analysis, and \textcolor{RoyalBlue}{$\mathtt{{<}characteristic{>}}$} is the specific aspect the LMM is tasked to focus on in the comparison. Recall that from each image sets pairs, we obtain multiple ComPairs samples, one for each characteristic.
\begin{table*}[ht]
\centering
\begin{tabular}{lp{13cm}}
\toprule
\multicolumn{2}{c}{
\begin{tabular}{m{0.0em} r m{0.0em} r}
    \rotatebox[origin=l]{90}{$\mathtt{{<}img1{>}}$}
    & 
    \makecell{
         \includegraphics[width=2.5cm]{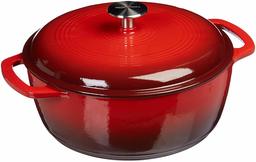}
         \includegraphics[width=2.5cm]{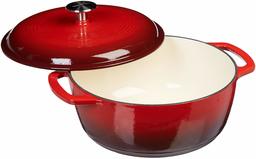} 
         \includegraphics[width=2.5cm]{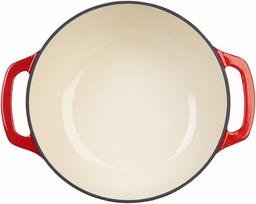}
         }
    &
    \rotatebox[origin=l]{90}{$\mathtt{{<}img2{>}}$}
    & 
    \makecell{
         \includegraphics[width=2.5cm]{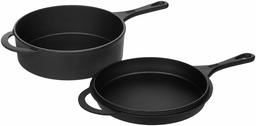}
         \includegraphics[width=2.5cm]{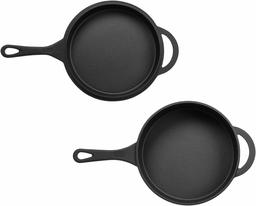} 
         \includegraphics[width=2.5cm]{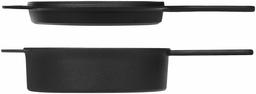}
         }
\end{tabular}}\\ 
\arrayrulecolor{gray}
\textbf{Prompt}: & Given the images of Product 1: $\mathtt{{<}img1{>}}$ and the images of Product 2: $\mathtt{{<}img2{>}}$, can you compare them in terms of \textcolor{RoyalBlue}{$\mathtt{{<}characteristic{>}}$}?\\ 
\arrayrulecolor{gray}\midrule
\textbf{Ground Truth}: & \\
\textcolor{RoyalBlue}{$\mathtt{{<}Handle{>}}$}: & Product 1 has loop side handles, Product 2 has long handle and an opposite helper handle on both pieces. \\
\textcolor{RoyalBlue}{$\mathtt{{<}Lid{>}}$}: & Product 1 has a secure-fitting lid, Product 2's skillet doubles as a secure-fitting lid. \\
\textcolor{RoyalBlue}{$\mathtt{{<}Color{>}}$}: & Product 1 is red, Product 2 does not have a specified color. \\
\arrayrulecolor{black}\midrule
\textbf{InternVL2-4B Vanilla}: & \\
\textcolor{RoyalBlue}{$\mathtt{{<}Handle{>}}$}: & The images of Product 1 and Product 2 show differences in the "Handle" as follows:

- Product 1 has a red handle, while Product 2 has a black handle. \\
\textcolor{RoyalBlue}{$\mathtt{{<}Lid{>}}$}: & The images of Product 1 and Product 2 show differences in the lid design. Product 1 has a red lid with a black handle and a black knob, while Product 2 has a black lid with a black handle and a black knob. \\
\textcolor{RoyalBlue}{$\mathtt{{<}Color{>}}$}: & The two products in the image differ in color. Product 1 is red, while Product 2 is black. \\
\arrayrulecolor{black}\midrule
\textbf{InternVL2-4B + \method}: & \\
\textcolor{RoyalBlue}{$\mathtt{{<}Handle{>}}$}: & The handles of Product 1 and Product 2 are different in color and material. Product 1 has a red handle, while Product 2 has a black handle. \\
\textcolor{RoyalBlue}{$\mathtt{{<}Lid{>}}$}: & The lid of Product 1 is red, while the lid of Product 2 is black. \\
\textcolor{RoyalBlue}{$\mathtt{{<}Color{>}}$}: & The two products have different colors: Product 1 is red, and Product 2 is black. \\
\bottomrule
\end{tabular}
\end{table*} \clearpage 
\begin{table*}[h!]
\centering
\begin{tabular}{lp{13cm}}
\toprule
\multicolumn{2}{c}{
\begin{tabular}{m{0.0em} r m{0.0em} r}
    \rotatebox[origin=l]{90}{$\mathtt{{<}img1{>}}$}
    & 
    \makecell{
         \includegraphics[width=2.5cm]{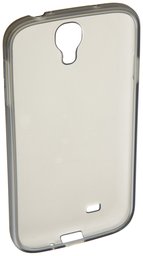}
         \includegraphics[width=2.5cm]{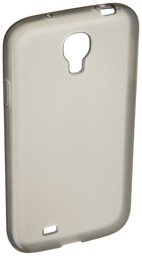} 
         \includegraphics[width=2.5cm]{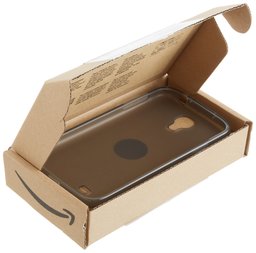}
         }
    &
    \rotatebox[origin=l]{90}{$\mathtt{{<}img2{>}}$}
    & 
    \makecell{
         \includegraphics[width=2.5cm]{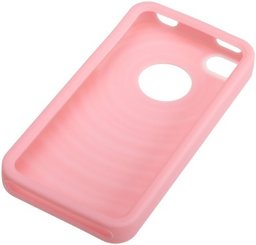}
         \includegraphics[width=2.5cm]{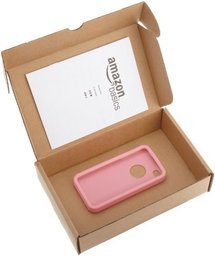} 
         \includegraphics[width=2.5cm]{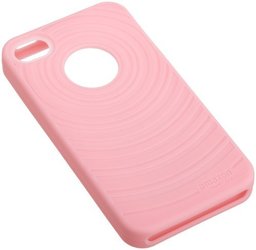}
         }
\end{tabular}}\\ 
\arrayrulecolor{gray}
\textbf{Prompt}: & Given the images of Product 1: $\mathtt{{<}img1{>}}$ and the images of Product 2: $\mathtt{{<}img2{>}}$, can you compare them in terms of \textcolor{RoyalBlue}{$\mathtt{{<}characteristic{>}}$}?\\ 
\arrayrulecolor{gray}\midrule
\textbf{Ground Truth}: & \\
\textcolor{RoyalBlue}{$\mathtt{{<}Color{>}}$}: & Product 1 is Smoke, while Product 2 is Pink. \\
\textcolor{RoyalBlue}{$\mathtt{{<}Dimensions{>}}$}: & Product 1 is slightly larger in length and width, but slightly thinner in height, compared to Product 2. \\
\textcolor{RoyalBlue}{$\mathtt{{<}Protection{>}}$}: & Both products offer protection, but Product 2's silicone case offers strong protection without adding bulk. \\
\textcolor{RoyalBlue}{$\mathtt{{<}Design{>}}$}: & Product 1 has a clear and flexible TPU material, while Product 2 has a soft-touch ripple pattern for improved grip. \\
\textcolor{RoyalBlue}{$\mathtt{{<}Material{>}}$}: & Product 1 is made of TPU, while Product 2 is made of Silicone. \\
\arrayrulecolor{black}\midrule
\textbf{InternVL2-4B Vanilla}: & \\
\textcolor{RoyalBlue}{$\mathtt{{<}Color{>}}$}: & The color of Product 1 is silver, while the color of Product 2 is pink. \\
\textcolor{RoyalBlue}{$\mathtt{{<}Dimensions{>}}$}: & The dimensions of Product 1 (the clear case) and Product 2 (the pink case) are different. Product 1 is larger and has a more rectangular shape, while Product 2 is smaller and has a more rounded shape. \\
\textcolor{RoyalBlue}{$\mathtt{{<}Protection{>}}$}: & The two products in the images are phone cases, and they differ in color and design. Product 1 is a clear case, while Product 2 is a pink case. Both cases have cutouts for the phone's camera and buttons, but the clear case allows the phone's design to be visible, whereas the pink case has a patterned design. \\
\textcolor{RoyalBlue}{$\mathtt{{<}Design{>}}$}: & The design of Product 1 is a clear, rectangular case with a single cutout for the camera and a smaller cutout for the flash. The case is plain and does not have any additional design elements or patterns.

The design of Product 2 is a pink, rectangular case with a wavy pattern on the surface. It also has a single cutout for the camera and a smaller cutout for the flash. The case has a more decorative and visually appealing design compared to Product 1. \\
\textcolor{RoyalBlue}{$\mathtt{{<}Material{>}}$}: & The material of Product 1 is clear plastic, while the material of Product 2 is pink plastic. \\
\arrayrulecolor{black}\midrule
\textbf{InternVL2-4B + \method}: & \\
\textcolor{RoyalBlue}{$\mathtt{{<}Color{>}}$}: & The two products have different colors. Product 1 is clear, while Product 2 is pink. \\
\textcolor{RoyalBlue}{$\mathtt{{<}Dimensions{>}}$}: & The dimensions of the two products appear to be different. Product 1 has a rectangular shape with rounded corners, while Product 2 has a more rounded rectangular shape with a curved edge. \\
\textcolor{RoyalBlue}{$\mathtt{{<}Protection{>}}$}: & The two products have different designs and materials, which may affect their protection levels. Product 1 appears to be made of a clear, possibly tempered glass material, which can provide good protection against scratches and minor impacts. Product 2, on the other hand, seems to be made of a plastic material with a textured surface, which might offer less protection against scratches and impacts compared to glass. \\
\textcolor{RoyalBlue}{$\mathtt{{<}Design{>}}$}: & The design of Product 1 is a clear, rectangular case with a single cutout for the camera, while Product 2 has a pink, textured design with a single cutout for the camera and a brand logo on the bottom right corner. \\
\textcolor{RoyalBlue}{$\mathtt{{<}Material{>}}$}: & The material of Product 1 appears to be a clear, possibly tempered glass or a hard plastic, while Product 2 seems to be made of a soft, flexible silicone or rubber. \\
\bottomrule
\end{tabular}
\end{table*} \clearpage 
\begin{table*}[h!]
\centering
\begin{tabular}{lp{13cm}}
\toprule
\multicolumn{2}{c}{
\begin{tabular}{m{0.0em} r m{0.0em} r}
    \rotatebox[origin=l]{90}{$\mathtt{{<}img1{>}}$}
    & 
    \makecell{
         \includegraphics[width=2.5cm]{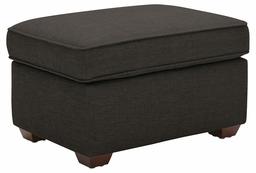}
         \includegraphics[width=2.5cm]{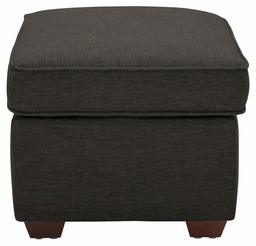} 
         \includegraphics[width=2.5cm]{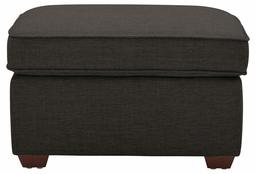}
         }
    &
    \rotatebox[origin=l]{90}{$\mathtt{{<}img2{>}}$}
    & 
    \makecell{
         \includegraphics[width=2.5cm]{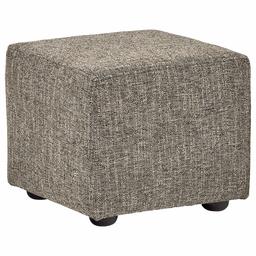}
         \includegraphics[width=2.5cm]{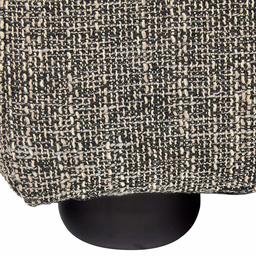} 
         \includegraphics[width=2.5cm]{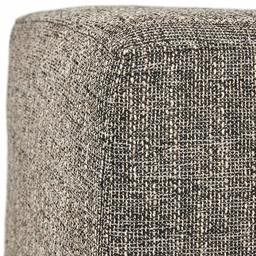}
         }
\end{tabular}}\\ 
\arrayrulecolor{gray}
\textbf{Prompt}: & Given the images of Product 1: $\mathtt{{<}img1{>}}$ and the images of Product 2: $\mathtt{{<}img2{>}}$, can you compare them in terms of \textcolor{RoyalBlue}{$\mathtt{{<}characteristic{>}}$}?\\ 
\arrayrulecolor{gray}\midrule
\textbf{Ground Truth}: & \\
\textcolor{RoyalBlue}{$\mathtt{{<}Color{>}}$}: & Product 1 is in Dark Grey, while Product 2 is in Boho (which is not a specific color, but a style/fabric combination). \\
\textcolor{RoyalBlue}{$\mathtt{{<}Fabric{>}}$}: & Product 2 is available in a range of fabrics, while Product 1's fabric is top-grain leather. \\
\textcolor{RoyalBlue}{$\mathtt{{<}Design{>}}$}: & Product 1 is a rectangular ottoman with a contemporary style, while Product 2 is a cube-shaped ottoman with a boho style. \\
\textcolor{RoyalBlue}{$\mathtt{{<}Material{>}}$}: & Product 1 has a solid hardwood frame, while Product 2 has a solid wood frame and polyester/acrylic blend fabric. \\
\arrayrulecolor{black}\midrule
\textbf{InternVL2-4B Vanilla}: & \\
\textcolor{RoyalBlue}{$\mathtt{{<}Color{>}}$}: & The two products in the images are identical in color. Both are dark gray or charcoal-colored cushioned ottomans with a textured fabric. \\
\textcolor{RoyalBlue}{$\mathtt{{<}Fabric{>}}$}: & The fabric of the two products appears to be the same. Both are upholstered in a dark gray, textured fabric with a subtle pattern. The texture and color consistency suggest that they are made from the same material. \\
\textcolor{RoyalBlue}{$\mathtt{{<}Design{>}}$}: & The two products in the images have the same design. They both feature a dark gray fabric upholstered on a square ottoman with a textured surface. The ottomans have a similar shape and size, with a slightly rounded top and a flat base. The fabric appears to be the same in both images, indicating that they are identical in design. \\
\textcolor{RoyalBlue}{$\mathtt{{<}Material{>}}$}: & The material of the couch in both images appears to be a textured fabric, likely a type of woven or knitted material. The texture suggests a durable and possibly high-quality fabric, suitable for furniture. The color is a dark, muted gray, which gives the couch a sophisticated and modern look. The couch has a simple design with a rounded top and four small, round wooden legs. \\
\arrayrulecolor{black}\midrule
\textbf{InternVL2-4B + \method}: & \\
\textcolor{RoyalBlue}{$\mathtt{{<}Color{>}}$}: & The color of both products appears to be the same. \\
\textcolor{RoyalBlue}{$\mathtt{{<}Fabric{>}}$}: & The fabric of Product 1 appears to be a smooth, dark gray material, while Product 2 has a more textured, possibly woven fabric with a mix of dark and light gray tones. \\
\textcolor{RoyalBlue}{$\mathtt{{<}Design{>}}$}: & The design of the two products is different in terms of the texture and pattern of the fabric. Product 1 has a smooth, solid-colored fabric, while Product 2 has a textured, patterned fabric. \\
\textcolor{RoyalBlue}{$\mathtt{{<}Material{>}}$}: & The material of Product 1 appears to be a smooth, possibly synthetic fabric, while Product 2 has a more textured, possibly woven fabric. \\
\bottomrule
\end{tabular}
\end{table*} \clearpage 
\begin{table*}[h!]
\centering
\begin{tabular}{lp{13cm}}
\toprule
\multicolumn{2}{c}{
\begin{tabular}{m{0.0em} r m{0.0em} r}
    \rotatebox[origin=l]{90}{$\mathtt{{<}img1{>}}$}
    & 
    \makecell{
         \includegraphics[width=2.5cm]{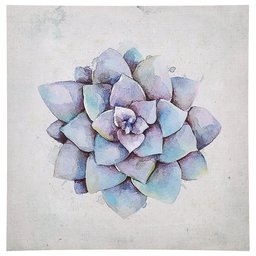}
         \includegraphics[width=2.5cm]{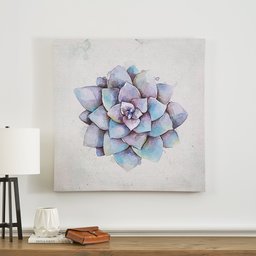} 
         \includegraphics[width=2.5cm]{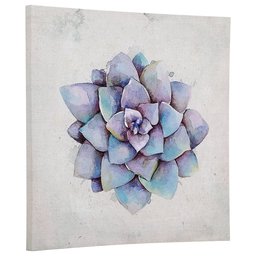}
         }
    &
    \rotatebox[origin=l]{90}{$\mathtt{{<}img2{>}}$}
    & 
    \makecell{
         \includegraphics[width=2.5cm]{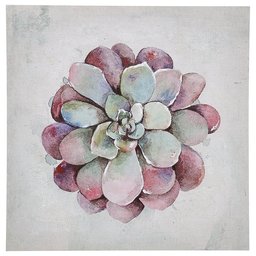}
         \includegraphics[width=2.5cm]{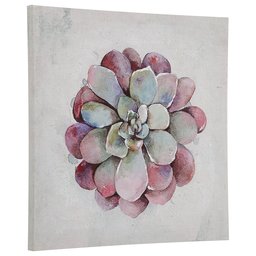} 
         \includegraphics[width=2.5cm]{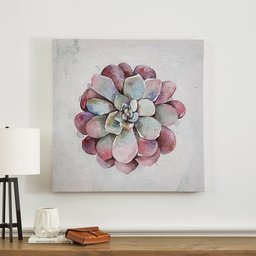}
         }
\end{tabular}}\\ 
\arrayrulecolor{gray}
\textbf{Prompt}: & Given the images of Product 1: $\mathtt{{<}img1{>}}$ and the images of Product 2: $\mathtt{{<}img2{>}}$, can you compare them in terms of \textcolor{RoyalBlue}{$\mathtt{{<}characteristic{>}}$}?\\ 
\arrayrulecolor{gray}\midrule
\textbf{Ground Truth}: & \\
\textcolor{RoyalBlue}{$\mathtt{{<}Color{>}}$}: & The colors in Product 1 are blue, purple, and green, whereas in Product 2, the colors are purple, pink, and green. \\
\textcolor{RoyalBlue}{$\mathtt{{<}Design{>}}$}: & The design of the two products is similar, with a floral art piece on a white background, but with different color combinations and flower types. \\
\textcolor{RoyalBlue}{$\mathtt{{<}Flower Type{>}}$}: & Product 1 features a succulent, while Product 2 is of an Echeveria succulent bloom. \\
\arrayrulecolor{black}\midrule
\textbf{InternVL2-4B Vanilla}: & \\
\textcolor{RoyalBlue}{$\mathtt{{<}Color{>}}$}: & The two products in the image have different color schemes. The first product features a watercolor painting of a succulent with shades of purple, blue, and green, while the second product has a similar watercolor painting but with shades of pink, red, and purple. \\
\textcolor{RoyalBlue}{$\mathtt{{<}Design{>}}$}: & The two products are similar in design, featuring a watercolor painting of a succulent plant. However, there are differences in the color palette and the specific shades used for the succulent leaves. The first product has a more varied color palette with shades of purple, blue, and green, while the second product has a more monochromatic palette with shades of pink and purple. \\
\textcolor{RoyalBlue}{$\mathtt{{<}Flower Type{>}}$}: & The two products in the image are similar in that they both feature a watercolor painting of a succulent plant. However, there are differences in the color palette and the specific type of succulent depicted. The first product shows a succulent with a more varied color palette, including shades of purple, blue, and green, while the second product features a succulent with a more monochromatic color scheme, primarily in shades of pink and purple. \\
\arrayrulecolor{black}\midrule
\textbf{InternVL2-4B + \method}: & \\
\textcolor{RoyalBlue}{$\mathtt{{<}Color{>}}$}: & The two products have different color schemes. Product 1 features a palette of soft blues and purples, while Product 2 has a combination of pinks and greens. \\
\textcolor{RoyalBlue}{$\mathtt{{<}Design{>}}$}: & The two products have different designs. Product 1 features a watercolor painting of succulents with a more abstract and blended style, while Product 2 has a more detailed and realistic depiction of succulents with distinct edges and shading. \\
\textcolor{RoyalBlue}{$\mathtt{{<}Flower Type{>}}$}: & The flower types in the two products are different. The first product features a succulent, while the second product features a cactus. \\
\bottomrule
\end{tabular}
\end{table*} \clearpage 
\begin{table*}[h!]
\centering
\begin{tabular}{lp{13cm}}
\toprule
\multicolumn{2}{c}{
\begin{tabular}{m{0.0em} r m{0.0em} r}
    \rotatebox[origin=l]{90}{$\mathtt{{<}img1{>}}$}
    & 
    \makecell{
         \includegraphics[width=2.5cm]{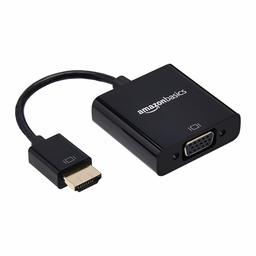}
         \includegraphics[width=2.5cm]{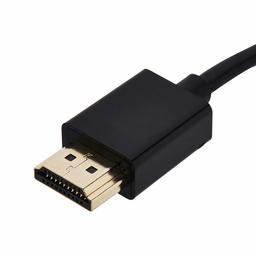} 
         \includegraphics[width=2.5cm]{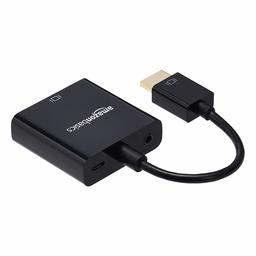}
         }
    &
    \rotatebox[origin=l]{90}{$\mathtt{{<}img2{>}}$}
    & 
    \makecell{
         \includegraphics[width=2.5cm]{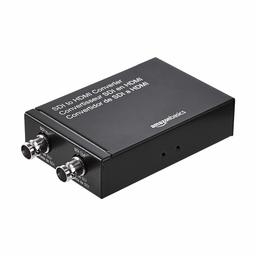}
         \includegraphics[width=2.5cm]{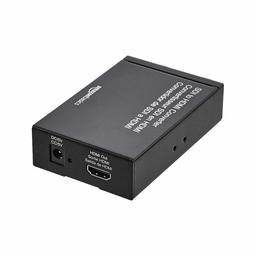} 
         \includegraphics[width=2.5cm]{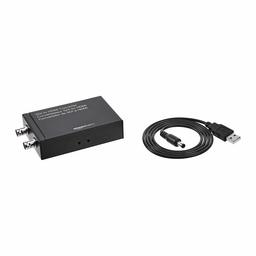}
         }
\end{tabular}}\\ 
\arrayrulecolor{gray}
\textbf{Prompt}: & Given the images of Product 1: $\mathtt{{<}img1{>}}$ and the images of Product 2: $\mathtt{{<}img2{>}}$, can you compare them in terms of \textcolor{RoyalBlue}{$\mathtt{{<}characteristic{>}}$}?\\ 
\arrayrulecolor{gray}\midrule
\textbf{Ground Truth}: & \\
\\
\textcolor{RoyalBlue}{$\mathtt{{<}Compatibility{>}}$}: & Product 1 is compatible with devices with standard HDMI ports, including desktop, laptop, ultrabook, and gaming consoles, while Product 2 is specifically designed for converting SDI signal to HDMI signal for professional video applications. \\
\textcolor{RoyalBlue}{$\mathtt{{<}Direction{>}}$}: & Product 1 is a one-way signal adapter, while Product 2 is a bi-directional converter, but with a separate input and output for SDI and HDMI. \\
\textcolor{RoyalBlue}{$\mathtt{{<}Purpose{>}}$}: & Product 1 is an HDMI to VGA adapter for presentations or events, while Product 2 is an SDI to HDMI converter for displaying media on an HDMI TV or monitor. \\
\arrayrulecolor{black}\midrule
\textbf{InternVL2-4B Vanilla}: & \\



\textcolor{RoyalBlue}{$\mathtt{{<}Compatibility{>}}$}: & The two products in the image are both SDI to HDMI converters made by AmazonBasics. The main difference between them is the cable type:

1. The first product on the left has a USB Type-A to HDMI cable.
2. The second product on the right has a standard SDI to HDMI cable.

Both products are designed to convert SDI (Serial Digital Interface) signals to HDMI (High-Definition Multimedia Interface) signals, allowing for the connection of SDI sources to HDMI displays. \hfill \xmark \\
\textcolor{RoyalBlue}{$\mathtt{{<}Direction{>}}$}: & The images show two different products from AmazonBasics. The first product is a USB to HDMI converter, and the second product is an HDMI to USB converter. Both products are designed to convert signals between USB and HDMI, allowing for the connection of USB devices to HDMI displays. \hfill \xmark \\



\textcolor{RoyalBlue}{$\mathtt{{<}Purpose{>}}$}: & The images show two different products from AmazonBasics. The first product is a USB to HDMI converter, and the second product is an HDMI to USB converter. Both products are designed to convert signals between USB and HDMI, allowing for the connection of USB devices to HDMI displays. \hfill \xmark \\
\arrayrulecolor{black}\midrule
\textbf{InternVL2-4B + \method}: & \\
\textcolor{RoyalBlue}{$\mathtt{{<}Compatibility{>}}$}: & The two products in the images are both AmazonBasics HDMI converters, but they differ in their compatibility. The product on the left is a standard HDMI converter, while the product on the right is an HDMI to VGA converter. \hfill \cmark \\
\textcolor{RoyalBlue}{$\mathtt{{<}Direction{>}}$}: & The images show two different USB-C to HDMI converters. The first converter has a USB-C port on one end and an HDMI port on the other, while the second converter has a USB-C port on one end and a USB-A port on the other. The first converter is designed for direct HDMI output, while the second converter can also connect to a computer via USB-A. \hfill \cmark \\



\textcolor{RoyalBlue}{$\mathtt{{<}Purpose{>}}$}: & The two products in the images are both AmazonBasics HDMI converters, but they differ in their specifications and intended use.

- **Product 1**: This is a standard HDMI converter that can convert SDI (Single-Link DVI) to HDMI. It is designed for converting the signal from a single-link DVI source to an HDMI output.

- **Product 2**: This is a more advanced HDMI converter that can convert SDI (Single-Link DVI) to HDMI and also has the capability to convert SDI to SDI. This means it can handle both single-link DVI to HDMI and SDI to SDI conversions, making it suitable for more complex setups.

In summary, the main difference in purpose between the two products is that Product 2 can handle both single-link DVI to HDMI and SDI to SDI conversions, while Product 1 is limited to single-link DVI to HDMI conversion. \hfill \cmark \\
\bottomrule
\end{tabular}
\end{table*} \clearpage 
\begin{table*}[h!]
\centering
\begin{tabular}{lp{13cm}}
\toprule
\multicolumn{2}{c}{
\begin{tabular}{m{0.0em} r m{0.0em} r}
    \rotatebox[origin=l]{90}{$\mathtt{{<}img1{>}}$}
    & 
    \makecell{
         \includegraphics[width=2.5cm]{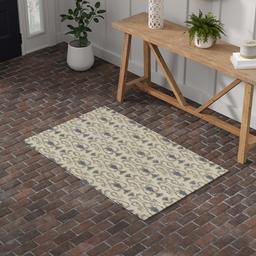}
         \includegraphics[width=2.5cm]{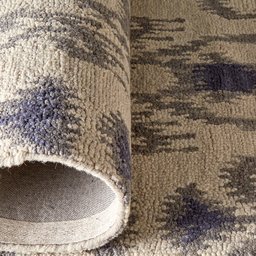} 
         \includegraphics[width=2.5cm]{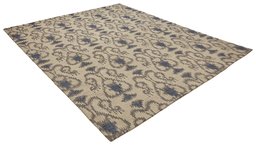}
         }
    &
    \rotatebox[origin=l]{90}{$\mathtt{{<}img2{>}}$}
    & 
    \makecell{
         \includegraphics[width=2.5cm]{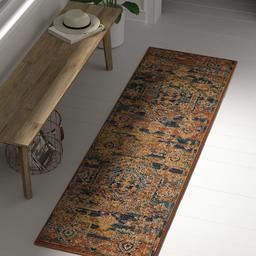}
         \includegraphics[width=2.5cm]{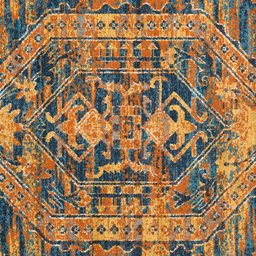} 
         \includegraphics[width=2.5cm]{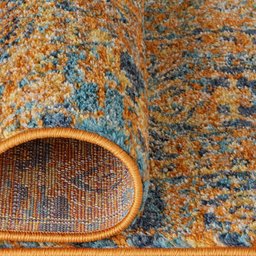}
         }
\end{tabular}}\\ 
\arrayrulecolor{gray}
\textbf{Prompt}: & Given the images of Product 1: $\mathtt{{<}img1{>}}$ and the images of Product 2: $\mathtt{{<}img2{>}}$, can you compare them in terms of \textcolor{RoyalBlue}{$\mathtt{{<}characteristic{>}}$}?\\ 
\arrayrulecolor{gray}\midrule
\textbf{Ground Truth}: & \\
\textcolor{RoyalBlue}{$\mathtt{{<}Material{>}}$}: & Product 1 is made of 100\% wool, whereas Product 2 is made of 100\% Polypropylene. \\
\textcolor{RoyalBlue}{$\mathtt{{<}Design{>}}$}: & Product 1 has a contemporary take on traditional design with an Ikat inspired pattern, whereas Product 2 has a traditional Persian rug with a strong geometric pattern. \\
\textcolor{RoyalBlue}{$\mathtt{{<}Pattern{>}}$}: & Product 1's pattern is described as'strikingly bold Ikat inspired' and 'neutral cool colors in over-sized motifs', whereas Product 2's pattern is described as'strong geometric pattern' with'sunny gold and rich teal blue colors'. \\
\arrayrulecolor{black}\midrule
\textbf{InternVL2-4B Vanilla}: & \\
\textcolor{RoyalBlue}{$\mathtt{{<}Material{>}}$}: & The material of the two products appears to be different. Product 1 is a rug with a textured, possibly woven or knitted fabric, while Product 2 is a flat, patterned fabric, likely a printed textile or a woven material. \\
\textcolor{RoyalBlue}{$\mathtt{{<}Design{>}}$}: & The design of the two products in the images is different. The first product has a more subtle, muted color palette with a beige background and dark blue patterns, while the second product features a vibrant, rich color scheme with a combination of orange, blue, and yellow hues. The first product has a more traditional, geometric pattern, whereas the second product has a more intricate, possibly floral or abstract design. \\


\textcolor{RoyalBlue}{$\mathtt{{<}Pattern{>}}$}: & The pattern on the two products is different. The first product has a beige background with dark blue and black geometric patterns, while the second product has a more complex and colorful pattern with a mix of orange, blue, and other colors. \\
\textcolor{RoyalBlue}{$\mathtt{{<}Care{>}}$}: & The two products in the images have different care instructions.

**Product 1:**
- The rug appears to be made of a natural fiber, likely wool or a wool blend.
- Care instructions for natural fiber rugs typically include:
  - Vacuuming regularly to remove dirt and dust.
  - Spot cleaning with a damp cloth for spills.
  - Avoiding harsh chemicals and detergents.
  - Dry cleaning or professional cleaning for deep stains.
  - Rotating the rug periodically to prevent uneven wear.

**Product 2:**
- The rug has a more intricate design and appears to be made of a synthetic material, possibly polypropylene or a blend.
- Care instructions for synthetic rugs typically include:
  - Vacuuming regularly to prevent dirt buildup.
  - Spot cleaning with a mild detergent and water.
  - Avoiding high-heat cleaning methods.
  - Professional cleaning for deep stains.
  - Rotating the rug periodically to prevent uneven wear.

In summary, while both rugs require regular vacuuming and spot cleaning, the natural fiber rug may need more delicate handling and professional cleaning, whereas the synthetic rug can be cleaned with mild detergents and is generally more durable. \\
\arrayrulecolor{black}\midrule
\textbf{InternVL2-4B + \method}: & \\
\textcolor{RoyalBlue}{$\mathtt{{<}Material{>}}$}: & The material of Product 1 appears to be a woven fabric, likely a type of rug or mat, while Product 2 seems to be a woven rug with a more intricate pattern. \\
\textcolor{RoyalBlue}{$\mathtt{{<}Design{>}}$}: & The design of the two products differs in terms of pattern complexity and color scheme. Product 1 has a simpler, more geometric pattern with a limited color palette, while Product 2 features a more intricate, abstract design with a broader range of colors. \\
\textcolor{RoyalBlue}{$\mathtt{{<}Pattern{>}}$}: & The pattern of the two products is different. Product 1 has a geometric pattern with a combination of beige, blue, and black colors, while Product 2 has a more organic, floral pattern with a combination of beige, blue, and black colors. \\
\bottomrule
\end{tabular}
\end{table*} \clearpage 
\begin{table*}[h!]
\centering
\begin{tabular}{lp{13cm}}
\toprule
\multicolumn{2}{c}{
\begin{tabular}{m{0.0em} r m{0.0em} r}
    \rotatebox[origin=l]{90}{$\mathtt{{<}img1{>}}$}
    & 
    \makecell{
         \includegraphics[width=2.5cm]{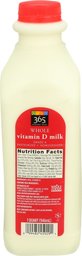}
         \includegraphics[width=2.5cm]{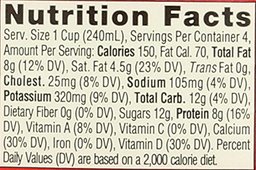} 
         \includegraphics[width=2.5cm]{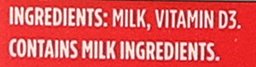}
         }
    &
    \rotatebox[origin=l]{90}{$\mathtt{{<}img2{>}}$}
    & 
    \makecell{
         \includegraphics[width=2.5cm]{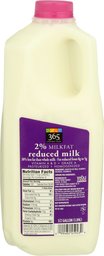}
         \includegraphics[width=2.5cm]{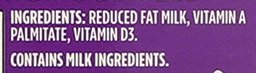} 
         \includegraphics[width=2.5cm]{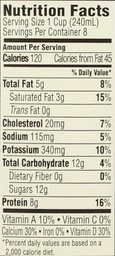}
         }
\end{tabular}}\\ 
\arrayrulecolor{gray}
\textbf{Prompt}: & Given the images of Product 1: $\mathtt{{<}img1{>}}$ and the images of Product 2: $\mathtt{{<}img2{>}}$, can you compare them in terms of \textcolor{RoyalBlue}{$\mathtt{{<}characteristic{>}}$}?\\ 
\arrayrulecolor{gray}\midrule
\textbf{Ground Truth}: & \\
\textcolor{RoyalBlue}{$\mathtt{{<}Item Volume{>}}$}: & Quart in Product 1, Half Gallon (64 oz) in Product 2 \\
\textcolor{RoyalBlue}{$\mathtt{{<}Milk Type{>}}$}: & Whole milk in Product 1, 2\% milkfat reduced milk in Product 2 \\
\textcolor{RoyalBlue}{$\mathtt{{<}Vitamins{>}}$}: & Vitamin D in Product 1, Vitamin A \& D in Product 2 \\
\textcolor{RoyalBlue}{$\mathtt{{<}Milk Fat Percentage{>}}$}: & Whole in Product 1, 2\% in Product 2 \\
\arrayrulecolor{black}\midrule
\textbf{InternVL2-4B Vanilla}: & \\
\textcolor{RoyalBlue}{$\mathtt{{<}Item Volume{>}}$}: & The two products differ in their "Item Volume" as the first product is a 1 quart (946 mL) container, while the second product is a 1/2 gallon (1.89 L) container. \\
\textcolor{RoyalBlue}{$\mathtt{{<}Milk Type{>}}$}: & The two products differ in their milk type: Product 1 is whole vitamin D milk, while Product 2 is reduced fat milk. \\
\textcolor{RoyalBlue}{$\mathtt{{<}Vitamins{>}}$}: & The two products differ in their vitamin content, with Product 1 containing Vitamin D and Product 2 containing Vitamin A and D3. \\
\textcolor{RoyalBlue}{$\mathtt{{<}Milk Fat Percentage{>}}$}: & The milk in Product 1 has a higher milk fat percentage than the milk in Product 2. \\
\arrayrulecolor{black}\midrule
\textbf{InternVL2-4B + \method}: & \\
\textcolor{RoyalBlue}{$\mathtt{{<}Item Volume{>}}$}: & The two products have different "Item Volumes." Product 1 is a 1 quart (948 mL) container, while Product 2 is a 1/2 gallon (1.89L) container. \\
\textcolor{RoyalBlue}{$\mathtt{{<}Milk Type{>}}$}: & The milk type in both products is "Whole Grade A." \\
\textcolor{RoyalBlue}{$\mathtt{{<}Vitamins{>}}$}: & The two products have different levels of Vitamin D. Product 1 has 12\% Vitamin D, while Product 2 has 0\% Vitamin D. \\
\textcolor{RoyalBlue}{$\mathtt{{<}Milk Fat Percentage{>}}$}: & The milk fat percentage in Product 1 is 4.5\%, while in Product 2 it is 2\%. \\
\bottomrule
\end{tabular}
\end{table*} \clearpage
\begin{table*}[h!]
\centering
\begin{tabular}{lp{13cm}}
\toprule
\multicolumn{2}{c}{
\begin{tabular}{m{0.0em} r m{0.0em} r}
    \rotatebox[origin=l]{90}{$\mathtt{{<}img1{>}}$}
    & 
    \makecell{
         \includegraphics[width=2.5cm]{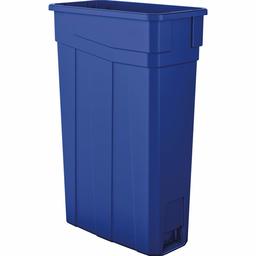}
         \includegraphics[width=2.5cm]{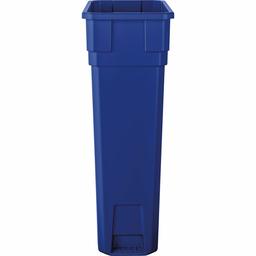} 
         \includegraphics[width=2.5cm]{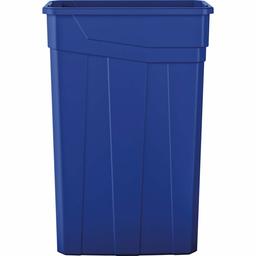}
         }
    &
    \rotatebox[origin=l]{90}{$\mathtt{{<}img2{>}}$}
    & 
    \makecell{
         \includegraphics[width=2.5cm]{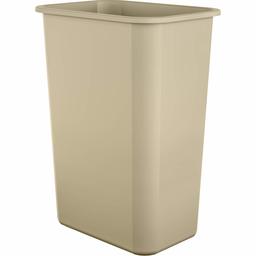}
         \includegraphics[width=2.5cm]{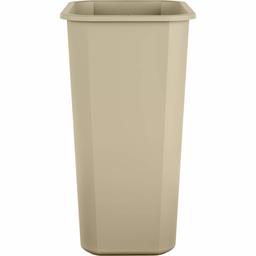} 
         \includegraphics[width=2.5cm]{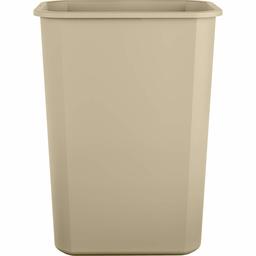}
         }
\end{tabular}}\\ 
\arrayrulecolor{gray}
\textbf{Prompt}: & Given the images of Product 1: $\mathtt{{<}img1{>}}$ and the images of Product 2: $\mathtt{{<}img2{>}}$, can you compare them in terms of \textcolor{RoyalBlue}{$\mathtt{{<}characteristic{>}}$}?\\ 
\arrayrulecolor{gray}\midrule
\textbf{Ground Truth}: & \\
\textcolor{RoyalBlue}{$\mathtt{{<}Material{>}}$}: & Both products are made of durable plastic, but Product 1 has a rugged construction for long-lasting strength and rigidity. \\
\textcolor{RoyalBlue}{$\mathtt{{<}Design{>}}$}: & Product 1 has a tall, slim profile with no side handles for out-of-the-way, space-saving placement, whereas Product 2 is a rectangular-shaped, space-efficient waste basket with a small footprint. \\
\textcolor{RoyalBlue}{$\mathtt{{<}Color{>}}$}: & Product 1 is Blue, while Product 2 is Beige. \\
\arrayrulecolor{black}\midrule
\textbf{InternVL2-4B Vanilla}: & \\
\textcolor{RoyalBlue}{$\mathtt{{<}Material{>}}$}: & The two products in the images are made of different materials. Product 1 is made of blue plastic, while Product 2 is made of beige plastic. \\
\textcolor{RoyalBlue}{$\mathtt{{<}Design{>}}$}: & The two products have different designs. Product 1 is blue with a slightly curved top and a handle on the side, while Product 2 is beige with a flat top and no visible handle. \\
\textcolor{RoyalBlue}{$\mathtt{{<}Color{>}}$}: & The two products in the image have different colors. Product 1 is blue, and Product 2 is beige. \\
\arrayrulecolor{black}\midrule
\textbf{InternVL2-4B + \method}: & \\
\textcolor{RoyalBlue}{$\mathtt{{<}Material{>}}$}: & The material of Product 1 appears to be a smooth, possibly plastic or coated metal, while Product 2 seems to be made of a matte, possibly plastic or coated metal as well. \\
\textcolor{RoyalBlue}{$\mathtt{{<}Design{>}}$}: & The design of Product 1 is blue, while the design of Product 2 is beige. \\
\textcolor{RoyalBlue}{$\mathtt{{<}Color{>}}$}: & The two products have different colors: Product 1 is blue, and Product 2 is beige. \\
\bottomrule
\end{tabular}
\end{table*} \clearpage
\begin{table*}[h!]
\centering
\begin{tabular}{lp{13cm}}
\toprule
\multicolumn{2}{c}{
\begin{tabular}{m{0.0em} r m{0.0em} r}
    \rotatebox[origin=l]{90}{$\mathtt{{<}img1{>}}$}
    & 
    \makecell{
         \includegraphics[width=2.5cm]{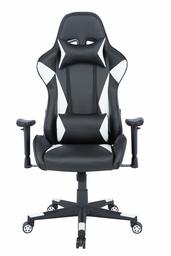}
         \includegraphics[width=2.5cm]{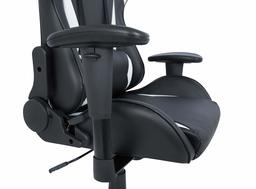} 
         \includegraphics[width=2.5cm]{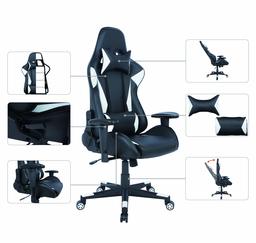}
         }
    &
    \rotatebox[origin=l]{90}{$\mathtt{{<}img2{>}}$}
    & 
    \makecell{
         \includegraphics[width=2.5cm]{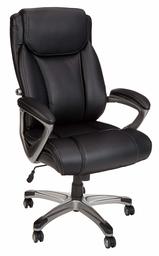}
         \includegraphics[width=2.5cm]{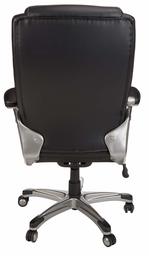} 
         \includegraphics[width=2.5cm]{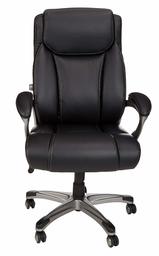}
         }
\end{tabular}}\\ 
\arrayrulecolor{gray}
\textbf{Prompt}: & Given the images of Product 1: $\mathtt{{<}img1{>}}$ and the images of Product 2: $\mathtt{{<}img2{>}}$, can you compare them in terms of \textcolor{RoyalBlue}{$\mathtt{{<}characteristic{>}}$}?\\ 
\arrayrulecolor{gray}\midrule
\textbf{Ground Truth}: & \\
\textcolor{RoyalBlue}{$\mathtt{{<}Swivel and Casters{>}}$}: & Both products have 360-degree swivel and smooth-rolling casters, but no specific difference is mentioned. \\
\textcolor{RoyalBlue}{$\mathtt{{<}Lumbar Support{>}}$}: & Product 1 has a lumbar cushion for added support, while Product 2 has an adjustable lumbar mechanism for lower-back support. \\
\textcolor{RoyalBlue}{$\mathtt{{<}Armrests{>}}$}: & Product 1 has custom fit with height-adjustable armrest, while Product 2 has curved arm rests. \\
\arrayrulecolor{black}\midrule
\textbf{InternVL2-4B Vanilla}: & \\
\textcolor{RoyalBlue}{$\mathtt{{<}Swivel and Casters{>}}$}: & The image shows a black office chair with a high backrest and armrests. The chair has a five-pointed star base with casters for mobility. The armrests are padded and have a slight curve to accommodate the user's arms. The backrest is ergonomically designed with contoured support for the lumbar region. The seat is also padded and appears to be made of leather or a leather-like material. The overall design suggests that the chair is intended for comfort and support during long periods of sitting, likely in an office setting. \\
\textcolor{RoyalBlue}{$\mathtt{{<}Lumbar Support{>}}$}: & The two products in the image are both black office chairs with a modern design. Here are the differences in their lumbar support:

1. **Product 1**:
   - The lumbar support is integrated into the backrest and is designed to provide targeted support to the lower back.
   - The lumbar support is shaped like a star, which may offer a more ergonomic fit for the lumbar region.
   - The overall design of the lumbar support appears to be more contoured and possibly adjustable.

2. **Product 2**:
   - The lumbar support is also integrated into the backrest but is more simplistic in design.
   - The lumbar support is less contoured compared to Product 1 and appears to be a single, flat piece.
   - The overall design is sleek and minimalistic, focusing on a streamlined look.

In summary, Product 1 offers a more contoured and potentially adjustable lumbar support, while Product 2 has a simpler, more minimalistic design. \\
\textcolor{RoyalBlue}{$\mathtt{{<}Armrests{>}}$}: & The armrests of the two chairs are different. The chair on the left has a more complex design with multiple armrests and a star-shaped design, while the chair on the right has a simpler, more traditional design with a single, rounded armrest. \\
\arrayrulecolor{black}\midrule
\textbf{InternVL2-4B + \method}: & \\
\textcolor{RoyalBlue}{$\mathtt{{<}Swivel and Casters{>}}$}: & Both products have swivel and casters, but the design and placement of the casters differ. Product 1 has casters on the base, while Product 2 has casters on the armrests. \\
\textcolor{RoyalBlue}{$\mathtt{{<}Lumbar Support{>}}$}: & The lumbar support in Product 1 appears to be more pronounced and contoured, with a clear emphasis on the lower back area. In contrast, Product 2 has a more subtle lumbar support design, with less visible contouring. \\
\textcolor{RoyalBlue}{$\mathtt{{<}Armrests{>}}$}: & The armrests of Product 1 and Product 2 are different. Product 1 has a more pronounced and angular design, while Product 2 has a smoother and more rounded armrest design. \\


\bottomrule
\end{tabular}
\end{table*} \clearpage 
\begin{table*}[h!]
\centering
\begin{tabular}{lp{13cm}}
\toprule
\multicolumn{2}{c}{
\begin{tabular}{m{0.0em} r m{0.0em} r}
    \rotatebox[origin=l]{90}{$\mathtt{{<}img1{>}}$}
    & 
    \makecell{
         \includegraphics[width=2.5cm]{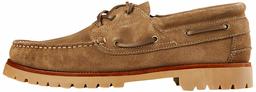}
         \includegraphics[width=2.5cm]{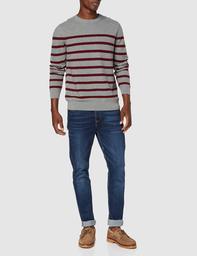} 
         \includegraphics[width=2.5cm]{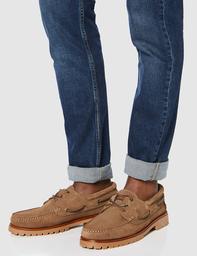}
         }
    &
    \rotatebox[origin=l]{90}{$\mathtt{{<}img2{>}}$}
    & 
    \makecell{
         \includegraphics[width=2.5cm]{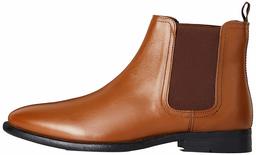}
         \includegraphics[width=2.5cm]{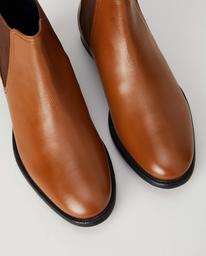} 
         \includegraphics[width=2.5cm]{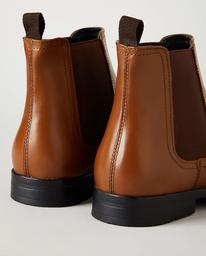}
         }
\end{tabular}}\\ 
\arrayrulecolor{gray}
\textbf{Prompt}: & Given the images of Product 1: $\mathtt{{<}img1{>}}$ and the images of Product 2: $\mathtt{{<}img2{>}}$, can you compare them in terms of \textcolor{RoyalBlue}{$\mathtt{{<}characteristic{>}}$}?\\ 
\arrayrulecolor{gray}\midrule
\textbf{Ground Truth}: & \\
\textcolor{RoyalBlue}{$\mathtt{{<}Color{>}}$}: & Product 1 has 'Beige Sand Suede', Product 2 has 'Brown Chelsea Tan' \\
\textcolor{RoyalBlue}{$\mathtt{{<}Style{>}}$}: & Product 1 doesn't have a specific style, Product 2 is 'Arbor' style \\
\arrayrulecolor{black}\midrule
\textbf{InternVL2-4B Vanilla}: & \\



\textcolor{RoyalBlue}{$\mathtt{{<}Color{>}}$}: & The two products in the image are both brown in color. However, there are differences in their design and structure.

1. **Product 1**:
   - Appears to be a pair of brown suede shoes.
   - Features a brown suede upper with visible stitching.
   - Has a brown leather sole with a textured tread pattern.
   - The design includes metal eyelets and a pull tab for ease of wearing.

2. **Product 2**:
   - Appears to be a pair of brown leather boots.
   - Features a smooth brown leather upper with visible stitching.
   - Has a black rubber sole with a textured tread pattern.
   - The design includes a pull tab and a zipper on the inner side for ease of wearing.

In summary, while both products are brown, they differ in the type of upper material (suede vs. leather) and the type of sole (leather vs. rubber). \\
\textcolor{RoyalBlue}{$\mathtt{{<}Style{>}}$}: & The two products in the image are different styles of shoes.

1. **Product 1**:
   - **Style**: This is a casual, suede-like shoe with a low-top design.
   - **Features**: It has a brown suede upper with visible stitching details, a brown leather sole with a wavy pattern, and a small pull tab on the back.
   - **Design**: The shoe has a casual, relaxed look with a slightly elevated heel and a rounded toe.

2. **Product 2**:
   - **Style**: This is a more formal, leather boot with a high-top design.
   - **Features**: It has a brown leather upper with visible stitching details, a black rubber sole, and a pull tab on the back.
   - **Design**: The boot has a more structured and formal appearance with a higher heel and a pointed toe.

In summary, the main difference in style between the two products is that Product 1 is casual and suede-like, while Product 2 is formal and leather-based. \\


\arrayrulecolor{black}\midrule
\textbf{InternVL2-4B + \method}: & \\
\textcolor{RoyalBlue}{$\mathtt{{<}Color{>}}$}: & The two products in the images have different colors. Product 1 is brown, while Product 2 is black. \\
\textcolor{RoyalBlue}{$\mathtt{{<}Style{>}}$}: & The two products have different styles:

- Product 1 has a more casual, rugged look with visible stitching and a chunky sole.
- Product 2 has a sleeker, more polished appearance with a smooth finish and a more streamlined sole. \\

\bottomrule
\end{tabular}
\end{table*} \clearpage 
\begin{table*}[h!]
\centering
\begin{tabular}{lp{13cm}}
\toprule
\multicolumn{2}{c}{
\begin{tabular}{m{0.0em} r m{0.0em} r}
    \rotatebox[origin=l]{90}{$\mathtt{{<}img1{>}}$}
    & 
    \makecell{
         \includegraphics[width=2.5cm]{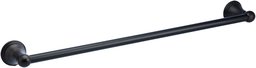}
         \includegraphics[width=2.5cm]{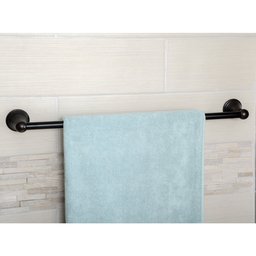} 
         \includegraphics[width=2.5cm]{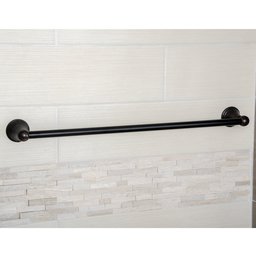}
         }
    &
    \rotatebox[origin=l]{90}{$\mathtt{{<}img2{>}}$}
    & 
    \makecell{
         \includegraphics[width=2.5cm]{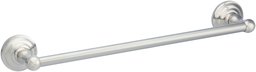}
         \includegraphics[width=2.5cm]{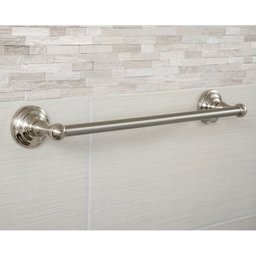} 
         \includegraphics[width=2.5cm]{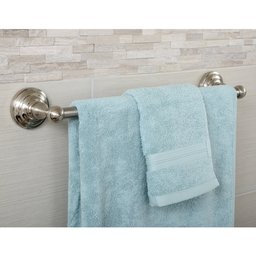}
         }
\end{tabular}}\\ 
\arrayrulecolor{gray}
\textbf{Prompt}: & Given the images of Product 1: $\mathtt{{<}img1{>}}$ and the images of Product 2: $\mathtt{{<}img2{>}}$, can you compare them in terms of \textcolor{RoyalBlue}{$\mathtt{{<}characteristic{>}}$}?\\ 
\arrayrulecolor{gray}\midrule
\textbf{Ground Truth}: & \\
\textcolor{RoyalBlue}{$\mathtt{{<}Color{>}}$}: & Product 1 has an Oil Rubbed Bronze finish while Product 2 has a Satin Nickel finish. \\
\textcolor{RoyalBlue}{$\mathtt{{<}Design{>}}$}: & Product 1 has a Modern design while Product 2 has a Traditional design. \\
\arrayrulecolor{black}\midrule
\textbf{InternVL2-4B Vanilla}: & \\
\textcolor{RoyalBlue}{$\mathtt{{<}Color{>}}$}: & The images show two different towel bars. The first towel bar is black, while the second towel bar is silver. \\



\textcolor{RoyalBlue}{$\mathtt{{<}Design{>}}$}: & The design of the towel bar in the image is sleek and modern, featuring a polished chrome finish that adds a touch of elegance to the bathroom. The bar is mounted on a light-colored wall with a textured tile backsplash, which complements the contemporary aesthetic. The towels hanging on the bar are neatly folded and appear to be of a soft, light blue color, adding a fresh and clean look to the space. The overall design is simple yet sophisticated, making it a stylish addition to any bathroom. \\
\arrayrulecolor{black}\midrule
\textbf{InternVL2-4B + \method}: & \\
\textcolor{RoyalBlue}{$\mathtt{{<}Color{>}}$}: & The color of the handles on both products is different: Product 1 has a dark color, while Product 2 has a silver color. \\
\textcolor{RoyalBlue}{$\mathtt{{<}Design{>}}$}: & The design of the two products differs in terms of finish and color. Product 1 has a matte black finish, while Product 2 has a shiny, metallic finish. \\
\bottomrule
\end{tabular}
\end{table*} \clearpage 
\begin{table*}[h!]
\centering
\begin{tabular}{lp{13cm}}
\toprule
\multicolumn{2}{c}{
\begin{tabular}{m{0.0em} r m{0.0em} r}
    \rotatebox[origin=l]{90}{$\mathtt{{<}img1{>}}$}
    & 
    \makecell{
         \includegraphics[width=2.5cm]{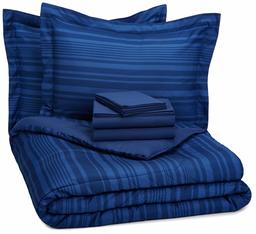}
         \includegraphics[width=2.5cm]{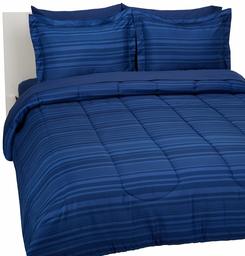} 
         \includegraphics[width=2.5cm]{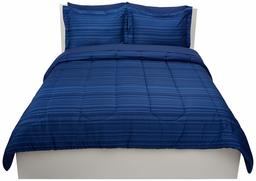}
         }
    &
    \rotatebox[origin=l]{90}{$\mathtt{{<}img2{>}}$}
    & 
    \makecell{
         \includegraphics[width=2.5cm]{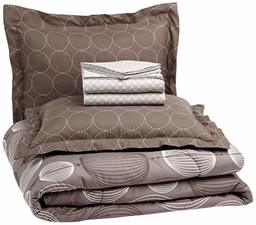}
         \includegraphics[width=2.5cm]{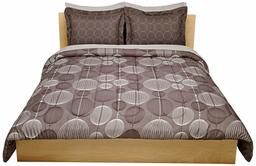} 
         \includegraphics[width=2.5cm]{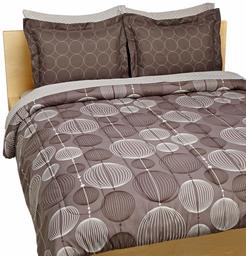}
         }
\end{tabular}}\\ 
\arrayrulecolor{gray}
\textbf{Prompt}: & Given the images of Product 1: $\mathtt{{<}img1{>}}$ and the images of Product 2: $\mathtt{{<}img2{>}}$, can you compare them in terms of \textcolor{RoyalBlue}{$\mathtt{{<}characteristic{>}}$}?\\ 
\arrayrulecolor{gray}\midrule
\textbf{Ground Truth}: & \\
\textcolor{RoyalBlue}{$\mathtt{{<}Color{>}}$}: & Product 1 has a 'Royal Blue Calvin Stripe' while Product 2 has an 'Industrial Grey'. \\
\textcolor{RoyalBlue}{$\mathtt{{<}Queen/King{>}}$}: & Both products are suitable for Queen-sized beds, but Product 1's 
not explicitly 
for King-sized beds, while Product 2's 
suitable for Queen and King-sized beds. \\
\textcolor{RoyalBlue}{$\mathtt{{<}Set Pieces{>}}$}: & Product 1 includes a 
comforter, 
flat sheet, 
fitted sheet, two 
shams, and two 
queen pillowcases, while Product 2 includes flat sheet, fitted sheet, 2 pillowcases, comforter, and 2 shams. \\
\arrayrulecolor{black}\midrule
\textbf{InternVL2-4B Vanilla}: & \\
\textcolor{RoyalBlue}{$\mathtt{{<}Color{>}}$}: & The two products in the images have different color schemes. Product 1 features a blue color scheme with blue and white stripes, while Product 2 has a brown color scheme with a pattern of circles and leaves. \\




\textcolor{RoyalBlue}{$\mathtt{{<}Queen/King{>}}$}: & The two products in the images are different in terms of their design and pattern.

- **Product 1** features a blue and white striped pattern.
- **Product 2** features a brown and white leaf pattern.

These differences in color and design make the two products distinct from each other. \\
\textcolor{RoyalBlue}{$\mathtt{{<}Set Pieces{>}}$}: & The set pieces in the two images are different in terms of their patterns and colors.

- **Product 1** features a blue and white striped pattern.
- **Product 2** features a brown and white circular pattern. \\
\arrayrulecolor{black}\midrule
\textbf{InternVL2-4B + \method}: & \\
\textcolor{RoyalBlue}{$\mathtt{{<}Color{>}}$}: & The two products have different colors. Product 1 is blue, while Product 2 is brown with a leaf pattern. \\
\textcolor{RoyalBlue}{$\mathtt{{<}Queen/King{>}}$}: & The two products in the images are different in terms of the "Queen/King" size. Product 1 is a Queen-sized bedding set, while Product 2 is a King-sized bedding set. \\
\textcolor{RoyalBlue}{$\mathtt{{<}Set Pieces{>}}$}: & The set pieces in the two products differ in color and pattern. Product 1 has a blue striped pattern, while Product 2 has a brown pattern with circular motifs. \\
\bottomrule
\end{tabular}
\end{table*} \clearpage 
\begin{table*}[h!]
\centering
\begin{tabular}{lp{13cm}}
\toprule
\multicolumn{2}{c}{
\begin{tabular}{m{0.0em} r m{0.0em} r}
    \rotatebox[origin=l]{90}{$\mathtt{{<}img1{>}}$}
    & 
    \makecell{
         \includegraphics[width=2.5cm]{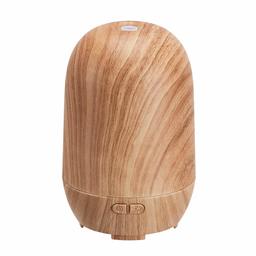}
         \includegraphics[width=2.5cm]{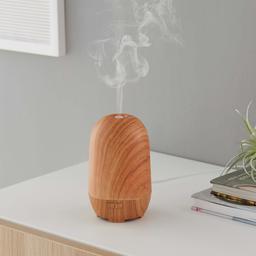} 
         \includegraphics[width=2.5cm]{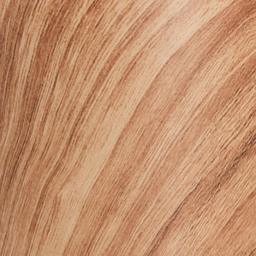}
         }
    &
    \rotatebox[origin=l]{90}{$\mathtt{{<}img2{>}}$}
    & 
    \makecell{
         \includegraphics[width=2.5cm]{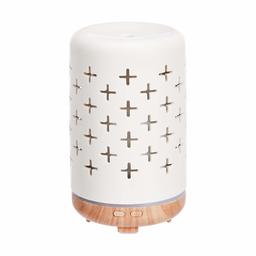}
         \includegraphics[width=2.5cm]{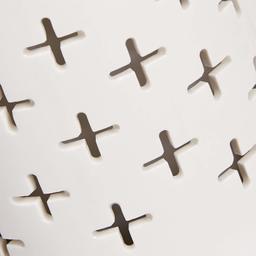} 
         \includegraphics[width=2.5cm]{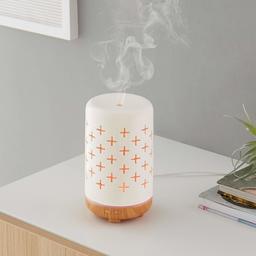}
         }
\end{tabular}}\\ 
\arrayrulecolor{gray}
\textbf{Prompt}: & Given the images of Product 1: $\mathtt{{<}img1{>}}$ and the images of Product 2: $\mathtt{{<}img2{>}}$, can you compare them in terms of \textcolor{RoyalBlue}{$\mathtt{{<}characteristic{>}}$}?\\ 
\arrayrulecolor{gray}\midrule
\textbf{Ground Truth}: & \\
\textcolor{RoyalBlue}{$\mathtt{{<}Design{>}}$}: & Product 1 has a classic wood grain finish, while Product 2 has a ceramic body with a hollow symmetric pattern design detailing. \\
\textcolor{RoyalBlue}{$\mathtt{{<}Style{>}}$}: & Product 1 is oval in shape, while Product 2 has a symmetric pattern design. \\
\textcolor{RoyalBlue}{$\mathtt{{<}Color of the Main Body{>}}$}: & Product 1's main body is classic wood grain, while Product 2's main body is ceramic with no specific color mentioned. \\
\textcolor{RoyalBlue}{$\mathtt{{<}Material of the Main Body{>}}$}: & Product 1's main body is made of wood, while Product 2's main body is made of ceramic. \\
\arrayrulecolor{black}\midrule
\textbf{InternVL2-4B Vanilla}: & \\
\textcolor{RoyalBlue}{$\mathtt{{<}Design{>}}$}: & The two products have different designs. Product 1 has a wooden base with a smooth, light-colored wood finish and a simple, minimalistic design. Product 2 has a white base with a pattern of crosses and a more intricate design. \\



\textcolor{RoyalBlue}{$\mathtt{{<}Style{>}}$}: & The two products have different styles. Product 1 has a wooden design with a natural wood finish and a simple, minimalistic appearance. Product 2 has a white, modern design with a pattern of crosses and a more industrial look. \\
\textcolor{RoyalBlue}{$\mathtt{{<}Color of the Main Body{>}}$}: & The color of the main body of Product 1 is light brown, while the color of the main body of Product 2 is white. \\
\textcolor{RoyalBlue}{$\mathtt{{<}Material of the Main Body{>}}$}: & The main body of the two products is made of different materials. Product 1 is made of wood, while Product 2 is made of plastic. \\
\arrayrulecolor{black}\midrule
\textbf{InternVL2-4B + \method}: & \\
\textcolor{RoyalBlue}{$\mathtt{{<}Design{>}}$}: & The design of the two products differs in terms of the pattern and placement of the cut-out holes on the surface. Product 1 has two symmetrical cut-out holes near the top, while Product 2 has multiple cut-out holes arranged in a grid pattern. \\
\textcolor{RoyalBlue}{$\mathtt{{<}Style{>}}$}: & The two products have different styles: Product 1 has a wooden texture with a simple design featuring two circular buttons, while Product 2 has a white, textured surface with a pattern of crosses. \\
\textcolor{RoyalBlue}{$\mathtt{{<}Color of the Main Body{>}}$}: & The color of the main body of Product 1 is light brown, while the color of the main body of Product 2 is white. \\
\textcolor{RoyalBlue}{$\mathtt{{<}Material of the Main Body{>}}$}: & The main body of Product 1 appears to be made of wood, while Product 2 has a white, possibly ceramic or plastic, body. \\
\bottomrule
\end{tabular}
\end{table*} \clearpage 
\begin{table*}[h!]
\centering
\begin{tabular}{lp{13cm}}
\toprule
\multicolumn{2}{c}{
\begin{tabular}{m{0.0em} r m{0.0em} r}
    \rotatebox[origin=l]{90}{$\mathtt{{<}img1{>}}$}
    & 
    \makecell{
         \includegraphics[width=2.5cm]{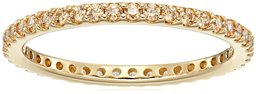}
         \includegraphics[width=2.5cm]{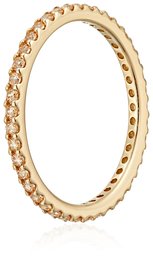} 
         \includegraphics[width=2.5cm]{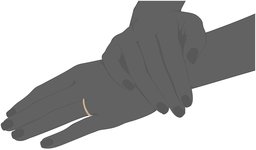}
         }
    &
    \rotatebox[origin=l]{90}{$\mathtt{{<}img2{>}}$}
    & 
    \makecell{
         \includegraphics[width=2.5cm]{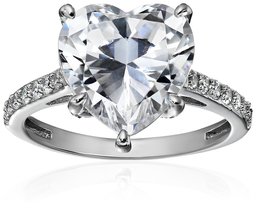}
         \includegraphics[width=2.5cm]{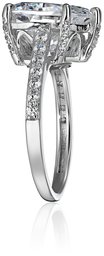} 
         \includegraphics[width=2.5cm]{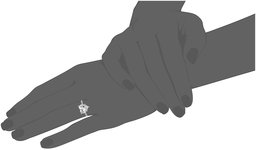}
         }
\end{tabular}}\\ 
\arrayrulecolor{gray}
\textbf{Prompt}: & Given the images of Product 1: $\mathtt{{<}img1{>}}$ and the images of Product 2: $\mathtt{{<}img2{>}}$, can you compare them in terms of \textcolor{RoyalBlue}{$\mathtt{{<}characteristic{>}}$}?\\ 
\arrayrulecolor{gray}\midrule
\textbf{Ground Truth}: & \\
\textcolor{RoyalBlue}{$\mathtt{{<}Color{>}}$}: & Product 1 is 'Yellow Gold Plated Silver' while Product 2 does not mention a specific color, only that it is 'Platinum-Plated Sterling Silver'. \\
\textcolor{RoyalBlue}{$\mathtt{{<}Design{>}}$}: & Product 1 is a 'Yellow Round-Cut All-Around Band Ring' while Product 2 is a 'Celebrity themed Engagement Ring' with a 'Heart-Shape Sparkler' \\
\arrayrulecolor{black}\midrule
\textbf{InternVL2-4B Vanilla}: & \\
\textcolor{RoyalBlue}{$\mathtt{{<}Color{>}}$}: & The images show two different styles of rings. 

1. The first ring has a large central diamond surrounded by smaller diamonds on the band. The band itself is adorned with smaller diamonds and features a unique design with a textured edge.
2. The second ring has a large central diamond with a halo of smaller diamonds around it. The band is also adorned with smaller diamonds and has a smooth, polished finish.

Both rings feature a prominent central diamond and are set with smaller diamonds on the band, but they differ in the design of the band and the overall aesthetic. \\
\textcolor{RoyalBlue}{$\mathtt{{<}Design{>}}$}: & The images of Product 1 and Product 2 show two different ring designs.

**Product 1:**
- The ring features a large central diamond set in a prong setting.
- The band is adorned with smaller diamonds along its sides.
- The band has a textured design with small, round, and pave-set diamonds.

**Product 2:**
- The ring also features a large central diamond set in a prong setting.
- The band is adorned with smaller diamonds along its sides.
- The band has a more intricate design with a combination of round and pave-set diamonds.

**Differences in Design:**
- Product 1 has a simpler band design with smaller, round, and pave-set diamonds.
- Product 2 has a more intricate band design with a combination of round and pave-set diamonds.

Both rings feature a large central diamond and smaller diamonds along the sides, but the band designs differ in complexity and the type of diamonds used. \\
\arrayrulecolor{black}\midrule
\textbf{InternVL2-4B + \method}: & \\
\textcolor{RoyalBlue}{$\mathtt{{<}Color{>}}$}: & The images of Product 1 and Product 2 both feature rings with a similar design, but there is a difference in the color of the stones. Product 1 has a gold-toned stone, while Product 2 has a clear, possibly diamond-colored stone. \\
\textcolor{RoyalBlue}{$\mathtt{{<}Design{>}}$}: & The design of Product 1 features a more intricate and textured band with multiple small diamonds, while Product 2 has a smoother, more polished band with a single large diamond. \\
\bottomrule
\end{tabular}
\end{table*} \clearpage 
\begin{table*}[h!]
\centering
\begin{tabular}{lp{13cm}}
\toprule
\multicolumn{2}{c}{
\begin{tabular}{m{0.0em} r m{0.0em} r}
    \rotatebox[origin=l]{90}{$\mathtt{{<}img1{>}}$}
    & 
    \makecell{
         \includegraphics[width=2.5cm]{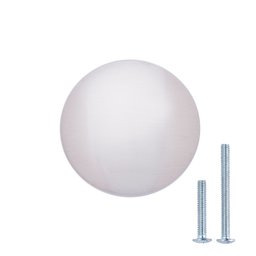}
         \includegraphics[width=2.5cm]{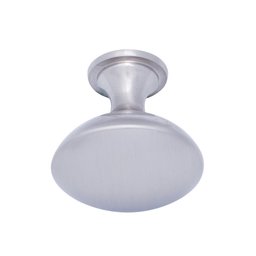} 
         \includegraphics[width=2.5cm]{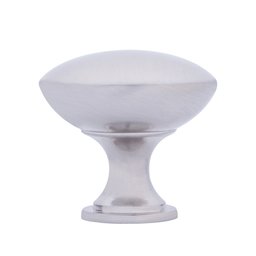}
         }
    &
    \rotatebox[origin=l]{90}{$\mathtt{{<}img2{>}}$}
    & 
    \makecell{
         \includegraphics[width=2.5cm]{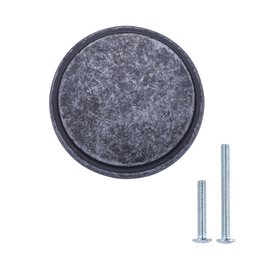}
         \includegraphics[width=2.5cm]{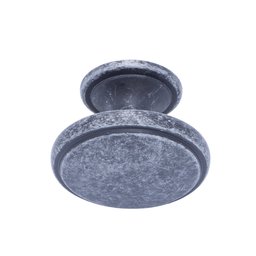} 
         \includegraphics[width=2.5cm]{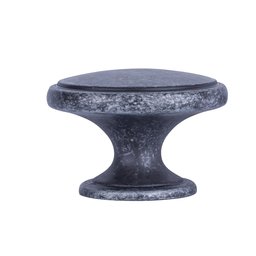}
         }
\end{tabular}}\\ 
\arrayrulecolor{gray}
\textbf{Prompt}: & Given the images of Product 1: $\mathtt{{<}img1{>}}$ and the images of Product 2: $\mathtt{{<}img2{>}}$, can you compare them in terms of \textcolor{RoyalBlue}{$\mathtt{{<}characteristic{>}}$}?\\ 
\arrayrulecolor{gray}\midrule
\textbf{Ground Truth}: & \\
\textcolor{RoyalBlue}{$\mathtt{{<}Finish{>}}$}: & Satin Nickel Finish in Product 1 and Antique Silver Finish in Product 2, with Satin Nickel being a darker, more neutral finish and Antique Silver being a more ornate, antique-inspired finish. \\
\arrayrulecolor{black}\midrule
\textbf{InternVL2-4B Vanilla}: & \\
\textcolor{RoyalBlue}{$\mathtt{{<}Finish{>}}$}: & The finish of the two products is different. Product 1 has a smooth, shiny finish, while Product 2 has a textured, matte finish. \\

\arrayrulecolor{black}\midrule
\textbf{InternVL2-4B + \method}: & \\
\textcolor{RoyalBlue}{$\mathtt{{<}Finish{>}}$}: & The finish of Product 1 is smooth and shiny, while the finish of Product 2 is matte and textured. \\
\bottomrule
\end{tabular}
\end{table*} \clearpage 
\begin{table*}[h!]
\centering
\begin{tabular}{lp{13cm}}
\toprule
\multicolumn{2}{c}{
\begin{tabular}{m{0.0em} r m{0.0em} r}
    \rotatebox[origin=l]{90}{$\mathtt{{<}img1{>}}$}
    & 
    \makecell{
         \includegraphics[width=2.5cm]{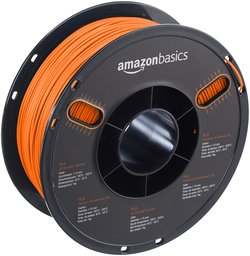}
         \includegraphics[width=2.5cm]{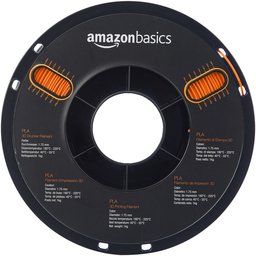} 
         \includegraphics[width=2.5cm]{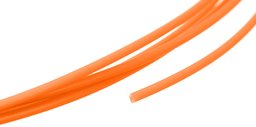}
         }
    &
    \rotatebox[origin=l]{90}{$\mathtt{{<}img2{>}}$}
    & 
    \makecell{
         \includegraphics[width=2.5cm]{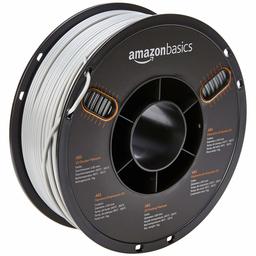}
         \includegraphics[width=2.5cm]{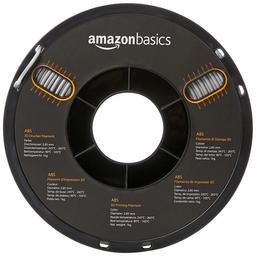} 
         \includegraphics[width=2.5cm]{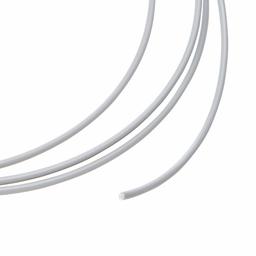}
         }
\end{tabular}}\\ 
\arrayrulecolor{gray}
\textbf{Prompt}: & Given the images of Product 1: $\mathtt{{<}img1{>}}$ and the images of Product 2: $\mathtt{{<}img2{>}}$, can you compare them in terms of \textcolor{RoyalBlue}{$\mathtt{{<}characteristic{>}}$}?\\ 
\arrayrulecolor{gray}\midrule
\textbf{Ground Truth}: & \\
\textcolor{RoyalBlue}{$\mathtt{{<}Color{>}}$}: & Product 1 is Orange, while Product 2 is Grey, differing in color. \\
\textcolor{RoyalBlue}{$\mathtt{{<}Material{>}}$}: & Product 1 is made of PLA, a commonly used thermoplastic material known for its lower melting temperature and ease of use, whereas Product 2 is made of ABS, a strong thermoplastic material great for creating durable items that require a higher resistance to heat. \\
\arrayrulecolor{black}\midrule
\textbf{InternVL2-4B Vanilla}: & \\
\textcolor{RoyalBlue}{$\mathtt{{<}Color{>}}$}: & The images show two different types of filament spools for 3D printers. The filament on the left is orange, while the filament on the right is white. \\


\textcolor{RoyalBlue}{$\mathtt{{<}Material{>}}$}: & The images show two different types of filament wires for 3D printers. The filament wires are labeled as "PLA" and "ABS."

1. **PLA Filament**:
   - Color: Orange
   - Diameter: 1.75 mm
   - Temperature range: 190°C - 220°C
   - Bed temperature: 40°C - 55°C
   - Net weight: 1 kg

2. **ABS Filament**:
   - Color: White
   - Diameter: 2.85 mm
   - Temperature range: 245°C - 265°C
   - Bed temperature: 90°C - 105°C
   - Net weight: 1 kg

The filament wires are coiled and have a black outer casing with orange and white text and markings. \\
\arrayrulecolor{black}\midrule
\textbf{InternVL2-4B + \method}: & \\
\textcolor{RoyalBlue}{$\mathtt{{<}Color{>}}$}: & The two products in the image are both 3D printer filaments, but they differ in color.

- **Product 1**: The filament is orange.
- **Product 2**: The filament is white.

These differences in color indicate that the filaments are intended for different printing projects or preferences. \\




\textcolor{RoyalBlue}{$\mathtt{{<}Material{>}}$}: & The two products in the image are both 3D printer filaments, but they differ in their material composition.

- **Product 1** is made of **PLA (Polylactic Acid)**.
- **Product 2** is made of **ABS (Acrylonitrile Butadiene Styrene)**.

PLA is a biodegradable thermoplastic derived from renewable resources like corn starch or sugarcane, while ABS is a petroleum-based plastic known for its strength and durability. \\



\bottomrule
\end{tabular}
\end{table*} \clearpage 
\begin{table*}[h!]
\centering
\begin{tabular}{lp{13cm}}
\toprule
\multicolumn{2}{c}{
\begin{tabular}{m{0.0em} r m{0.0em} r}
    \rotatebox[origin=l]{90}{$\mathtt{{<}img1{>}}$}
    & 
    \makecell{
         \includegraphics[width=2.5cm]{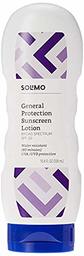}
         \includegraphics[width=2.5cm]{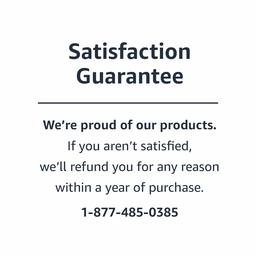} 
         \includegraphics[width=2.5cm]{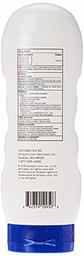}
         }
    &
    \rotatebox[origin=l]{90}{$\mathtt{{<}img2{>}}$}
    & 
    \makecell{
         \includegraphics[width=2.5cm]{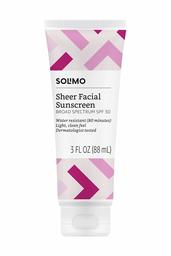}
         \includegraphics[width=2.5cm]{images/abo_imgs/20533b83.jpg} 
         \includegraphics[width=2.5cm]{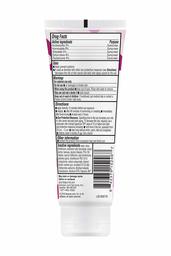}
         }
\end{tabular}}\\ 
\arrayrulecolor{gray}
\textbf{Prompt}: & Given the images of Product 1: $\mathtt{{<}img1{>}}$ and the images of Product 2: $\mathtt{{<}img2{>}}$, can you compare them in terms of \textcolor{RoyalBlue}{$\mathtt{{<}characteristic{>}}$}?\\ 
\arrayrulecolor{gray}\midrule
\textbf{Ground Truth}: & \\
\textcolor{RoyalBlue}{$\mathtt{{<}SPF{>}}$}: & Product 1 has SPF 50, while Product 2 has SPF 30. \\
\textcolor{RoyalBlue}{$\mathtt{{<}Volume of Product{>}}$}: & Product 1 is larger, with a volume of 10.4 Fl Oz, while Product 2 is smaller, with a volume of 3.0 fluid ounces. \\
\arrayrulecolor{black}\midrule
\textbf{InternVL2-4B Vanilla}: & \\
\textcolor{RoyalBlue}{$\mathtt{{<}SPF{>}}$}: & The two products have different SPF ratings. Product 1 has an SPF of 50, while Product 2 has an SPF of 30. \\
\textcolor{RoyalBlue}{$\mathtt{{<}Volume of Product{>}}$}: & The two products have different volumes. Product 1 has a volume of 10.4 FL OZ (300 mL), while Product 2 has a volume of 3 FL OZ (88 mL). \\
\arrayrulecolor{black}\midrule
\textbf{InternVL2-4B + \method}: & \\
\textcolor{RoyalBlue}{$\mathtt{{<}SPF{>}}$}: & The two products have different SPF levels. Product 1 has an SPF of 50, while Product 2 has an SPF of 30. \\
\textcolor{RoyalBlue}{$\mathtt{{<}Volume of Product{>}}$}: & The two products have different SPF levels. Product 1 has an SPF of 50, while Product 2 has an SPF of 30. \\
\bottomrule
\end{tabular}
\end{table*} \clearpage 
\begin{table*}[h!]
\centering
\begin{tabular}{lp{13cm}}
\toprule
\multicolumn{2}{c}{
\begin{tabular}{m{0.0em} r m{0.0em} r}
    \rotatebox[origin=l]{90}{$\mathtt{{<}img1{>}}$}
    & 
    \makecell{
         \includegraphics[width=2.5cm]{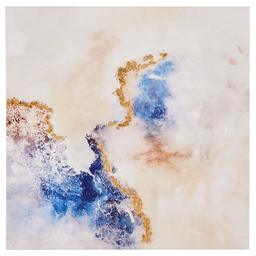}
         \includegraphics[width=2.5cm]{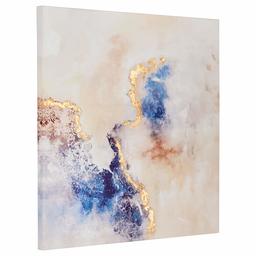} 
         \includegraphics[width=2.5cm]{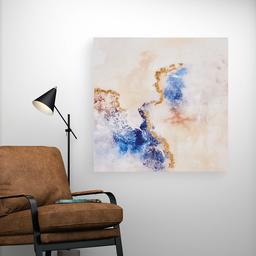}
         }
    &
    \rotatebox[origin=l]{90}{$\mathtt{{<}img2{>}}$}
    & 
    \makecell{
         \includegraphics[width=2.5cm]{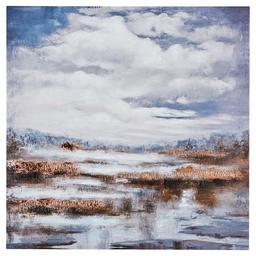}
         \includegraphics[width=2.5cm]{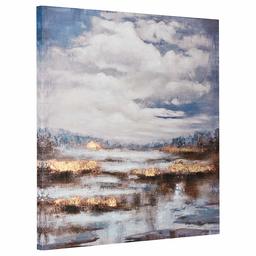} 
         \includegraphics[width=2.5cm]{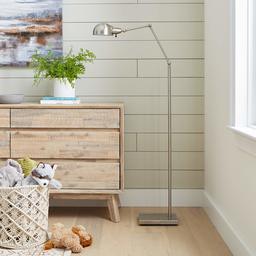}
         }
\end{tabular}}\\ 
\arrayrulecolor{gray}
\textbf{Prompt}: & Given the images of Product 1: $\mathtt{{<}img1{>}}$ and the images of Product 2: $\mathtt{{<}img2{>}}$, can you compare them in terms of \textcolor{RoyalBlue}{$\mathtt{{<}characteristic{>}}$}?\\ 
\arrayrulecolor{gray}\midrule
\textbf{Ground Truth}: & \\
\textcolor{RoyalBlue}{$\mathtt{{<}Design{>}}$}: & Product 1 has an abstract design, whereas Product 2 is a landscape print. \\
\textcolor{RoyalBlue}{$\mathtt{{<}Style{>}}$}: & Product 1 is Contemporary, whereas Product 2 is Traditional. \\
\textcolor{RoyalBlue}{$\mathtt{{<}Color Palette{>}}$}: & Product 1 has a unique color palette of blue, white, and beige, whereas Product 2 features shades of blue, white clouds, and copper leaf. \\
\arrayrulecolor{black}\midrule
\textbf{InternVL2-4B Vanilla}: & \\




\textcolor{RoyalBlue}{$\mathtt{{<}Design{>}}$}: & The two products in the images have different designs. The first product features a more abstract and colorful design with a mix of blue, gold, and white hues, while the second product has a more realistic and naturalistic design depicting a landscape with clouds and a body of water. \\
\textcolor{RoyalBlue}{$\mathtt{{<}Style{>}}$}: & The two products in the image have different styles. The first product features a more abstract and colorful design with a mix of blue, gold, and beige tones, creating a vibrant and dynamic appearance. The second product has a more realistic and serene depiction of a landscape with a focus on natural elements like water, clouds, and trees, using a palette of blues, whites, and earthy tones. \\
\textcolor{RoyalBlue}{$\mathtt{{<}Color Palette{>}}$}: & The two products have different color palettes. Product 1 features a mix of blue, gold, and beige tones, while Product 2 predominantly uses shades of blue and white. \\



\arrayrulecolor{black}\midrule
\textbf{InternVL2-4B + \method}: & \\
\textcolor{RoyalBlue}{$\mathtt{{<}Design{>}}$}: & The two products have different designs. Product 1 has a more abstract and colorful design with a mix of blue, gold, and white, while Product 2 has a more realistic and natural design depicting a landscape scene with clouds and a body of water. \\
\textcolor{RoyalBlue}{$\mathtt{{<}Style{>}}$}: & The two products have different styles. Product 1 has a more abstract and modern style with a mix of colors and textures, while Product 2 has a more realistic and natural style depicting a landscape scene. \\
\textcolor{RoyalBlue}{$\mathtt{{<}Color Palette{>}}$}: & The color palettes of the two products are different. Product 1 has a more muted and earthy palette with shades of blue, gold, and white, while Product 2 has a cooler and more subdued palette with shades of blue, white, and brown. \\
\bottomrule
\end{tabular}
\end{table*} \clearpage 
\begin{table*}[h!]
\centering
\begin{tabular}{lp{13cm}}
\toprule
\multicolumn{2}{c}{
\begin{tabular}{m{0.0em} r m{0.0em} r}
    \rotatebox[origin=l]{90}{$\mathtt{{<}img1{>}}$}
    & 
    \makecell{
         \includegraphics[width=2.5cm]{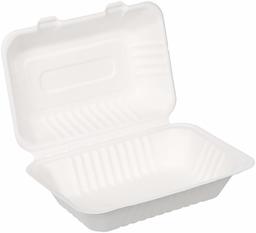}
         \includegraphics[width=2.5cm]{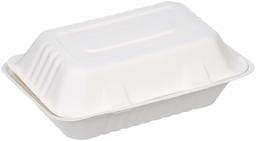} 
         \includegraphics[width=2.5cm]{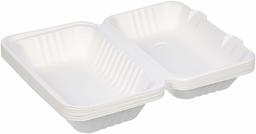}
         }
    &
    \rotatebox[origin=l]{90}{$\mathtt{{<}img2{>}}$}
    & 
    \makecell{
         \includegraphics[width=2.5cm]{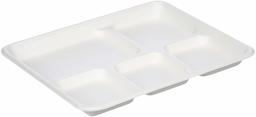}
         \includegraphics[width=2.5cm]{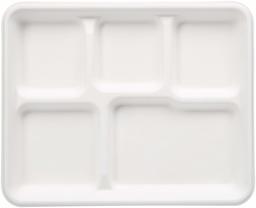} 
         \includegraphics[width=2.5cm]{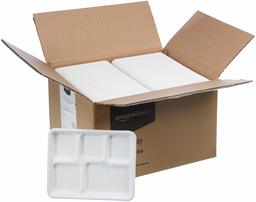}
         }
\end{tabular}}\\ 
\arrayrulecolor{gray}
\textbf{Prompt}: & Given the images of Product 1: $\mathtt{{<}img1{>}}$ and the images of Product 2: $\mathtt{{<}img2{>}}$, can you compare them in terms of \textcolor{RoyalBlue}{$\mathtt{{<}characteristic{>}}$}?\\ 
\arrayrulecolor{gray}\midrule
\textbf{Ground Truth}: & \\
\textcolor{RoyalBlue}{$\mathtt{{<}Color{>}}$}: & Product 1 is in a natural color, while Product 2 is white. \\
\textcolor{RoyalBlue}{$\mathtt{{<}Design{>}}$}: & Product 1 is a clamshell hinged food container, while Product 2 is a 5-compartment food tray. \\
\arrayrulecolor{black}\midrule
\textbf{InternVL2-4B Vanilla}: & \\
\textcolor{RoyalBlue}{$\mathtt{{<}Color{>}}$}: & The images show two white plastic trays. Both trays are identical in shape and size, with a grid pattern of raised ridges on the surface. The trays are designed to hold food items and are likely used for serving or storing food. The trays are placed inside a cardboard box, which is labeled with the AmazonBasics logo, indicating that they are likely a product from Amazon's AmazonBasics brand. The trays are empty and appear to be new. \\



\textcolor{RoyalBlue}{$\mathtt{{<}Design{>}}$}: & The two products in the image are both white, rectangular, and appear to be made of plastic. However, there are differences in their design:

1. **Product 1**:
   - The design is a single, large rectangular piece with a slightly raised, ridged edge.
   - It has a single compartment.

2. **Product 2**:
   - The design is a larger rectangular piece divided into four smaller compartments.
   - Each compartment is square and has a slightly raised, ridged edge.

In summary, Product 1 is a single, larger compartment, while Product 2 is divided into four smaller compartments. \\
\arrayrulecolor{black}\midrule
\textbf{InternVL2-4B + \method}: & \\
\textcolor{RoyalBlue}{$\mathtt{{<}Color{>}}$}: & The color of both products is white. \\
\textcolor{RoyalBlue}{$\mathtt{{<}Design{>}}$}: & The design of Product 1 and Product 2 is different. Product 1 has a smooth, untextured surface, while Product 2 has a textured surface with raised ridges. \\
\bottomrule
\end{tabular}
\end{table*} \clearpage 
\begin{table*}[h!]
\centering
\begin{tabular}{lp{13cm}}
\toprule
\multicolumn{2}{c}{
\begin{tabular}{m{0.0em} r m{0.0em} r}
    \rotatebox[origin=l]{90}{$\mathtt{{<}img1{>}}$}
    & 
    \makecell{
         \includegraphics[width=2.5cm]{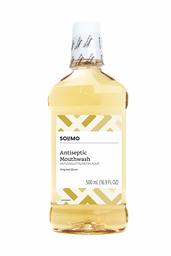}
         \includegraphics[width=2.5cm]{images/abo_imgs/20533b83.jpg} 
         \includegraphics[width=2.5cm]{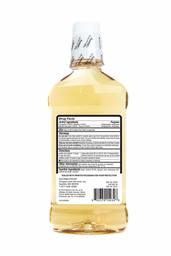}
         }
    &
    \rotatebox[origin=l]{90}{$\mathtt{{<}img2{>}}$}
    & 
    \makecell{
         \includegraphics[width=2.5cm]{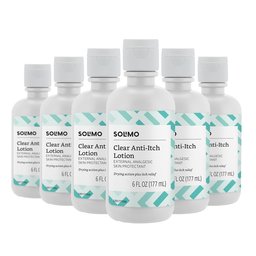}
         \includegraphics[width=2.5cm]{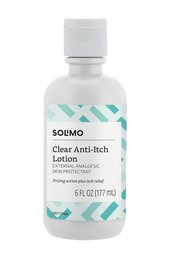} 
         \includegraphics[width=2.5cm]{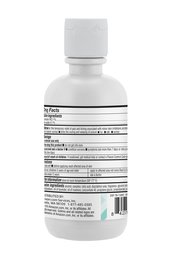}
         }
\end{tabular}}\\ 
\arrayrulecolor{gray}
\textbf{Prompt}: & Given the images of Product 1: $\mathtt{{<}img1{>}}$ and the images of Product 2: $\mathtt{{<}img2{>}}$, can you compare them in terms of \textcolor{RoyalBlue}{$\mathtt{{<}characteristic{>}}$}?\\ 
\arrayrulecolor{gray}\midrule
\textbf{Ground Truth}: & \\
\textcolor{RoyalBlue}{$\mathtt{{<}Product Purpose{>}}$}: & Product 1 is an antiseptic mouthwash for oral care, whereas Product 2 is an anti-itch lotion for skin care. \\
\textcolor{RoyalBlue}{$\mathtt{{<}Product Size{>}}$}: & Product 1 is a single 16.9-fluid ounce bottle, whereas Product 2 is a pack of six 6-fluid ounce bottles. \\
\arrayrulecolor{black}\midrule
\textbf{InternVL2-4B Vanilla}: & \\


\textcolor{RoyalBlue}{$\mathtt{{<}Product Purpose{>}}$}: & The product in the image is a "Solimo Clear Anti-Ich Lotion." It is an external analgesic skin protectant that provides drying action plus itch relief. The lotion is designed to help with itching associated with minor skin irritations and rashes. The active ingredients include Hydrochloric Acid (HCl) and Iodine (I2). The lotion is suitable for use on all skin types and is not intended for use in the eyes. It should be applied to the affected area of skin, and if symptoms persist for more than 7 days or if the condition worsens, medical help should be sought. The lotion is distributed by Amazon.com Services, Inc. \\

\textcolor{RoyalBlue}{$\mathtt{{<}Product Size{>}}$}: & The product sizes of the two products are different. The first product is a 500 mL (16.9 FL OZ) bottle, while the second product is a 6 FL OZ (177 mL) bottle. \\


\arrayrulecolor{black}\midrule
\textbf{InternVL2-4B + \method}: & \\



\textcolor{RoyalBlue}{$\mathtt{{<}Product Purpose{>}}$}: & The two products have different purposes:

- **Product 1** is an **Antiseptic Mouthwash**. It is designed to help control plaque and gingivitis, which are common oral health issues. The product is intended to be used as part of a daily oral hygiene routine to help maintain healthy gums and teeth.

- **Product 2** is a **Clear Anti-Ich Lotion**. This product is intended to provide relief from itching caused by external allergic reactions, such as those from poison ivy, poison oak, or poison sumac. It is designed to be applied to the affected area to soothe and reduce discomfort from itching. \\

\textcolor{RoyalBlue}{$\mathtt{{<}Product Size{>}}$}: & The two products have different sizes. The product on the left is a 500 mL (16.9 FL OZ) bottle, while the product on the right is a 6 FL OZ (177 mL) bottle. \\
\bottomrule
\end{tabular}
\end{table*} \clearpage 


\end{document}